\documentclass{article}
\usepackage[a4paper]{geometry}
\usepackage{lineno,hyperref}
\usepackage{graphicx}	
\usepackage[table]{xcolor}	
\usepackage{float}
\usepackage{subcaption}
\usepackage{amssymb}
\usepackage{amsmath}
\usepackage{mathtools}
\usepackage[ruled,vlined,linesnumbered]{algorithm2e}
\usepackage{makecell}
\newcommand{\citep}[1]{\cite{#1}}

\title{De-homogenization using Convolutional Neural Networks}
\author{Martin O. Elingaard$^{a,b}$, Niels Aage$^a$, J. Andreas Bærentzen$^b$, Ole Sigmund$^a$\\
 	{\small $^a$\textit{Dept. of Mechanical Engineering}, $^b$\textit{Dept. of Applied Mathematics and Computer Science}}\\
 	{\small \textit{Technical University of Denmark, Kgs. Lyngby, Denmark}}
}	
\date{\today}
\begin{document}
\maketitle
\begin{abstract}
This paper presents a deep learning-based de-homogenization method for structural compliance minimization. By using a convolutional neural network to parameterize the mapping from a set of lamination parameters on a coarse mesh to a one-scale design on a fine mesh, we avoid solving the least square problems associated with traditional de-homogenization approaches and save time correspondingly. To train the neural network, a two-step custom loss function has been developed which ensures a periodic output field that follows the local lamination orientations. A key feature of the proposed method is that the training is carried out without any use of or reference to the underlying structural optimization problem, which renders the proposed method robust and insensitive wrt. domain size, boundary conditions, and loading. A post-processing procedure utilizing a distance transform on the output field skeleton is used to project the desired lamination widths onto the output field while ensuring a predefined minimum length-scale and volume fraction. To demonstrate that the deep learning approach has excellent generalization properties, numerical examples are shown for several different load and boundary conditions. For an appropriate choice of parameters, the de-homogenized designs perform within $7-25\%$ of the homogenization-based solution at a fraction of the computational cost. With several options for further improvements, the scheme may provide the basis for future interactive high-resolution topology optimization.
\end{abstract}



\begin{figure}[H]
	\centering
	\includegraphics[width=0.90\textwidth]{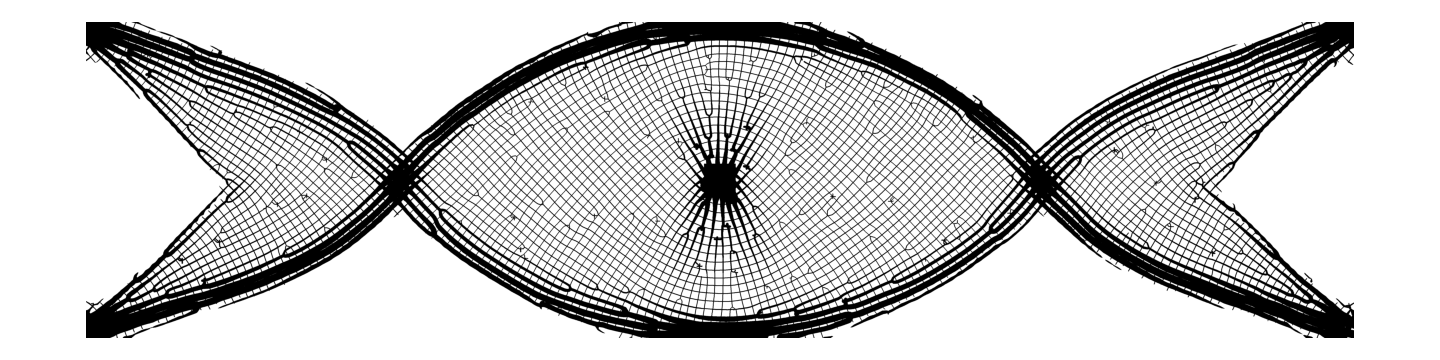}
	\caption{De-homogenization of a $200\times50$ homogenization-based solution onto a fine mesh of $8000\times2000$ elements. Compliance of the de-homogenized design is within $17\%$ of the homogenization solution, adheres to a predefined minimum relative thickness, and has been de-homogenized in just 15 seconds on a modern day laptop. }
	\label{fig:eye_catcher}
\end{figure}

\section{Introduction}

In many ways the fields of computer vision and topology optimization are closely related, e.g. in both cases, the optimization domain constitutes a set of elements/pixels in a grid, and the problem can be formulated as minimizing a functional to obtain an optimized configuration. In computer vision, an optimized configuration could be a set of segmentation labels, which in the context of topology optimization \cite{bendsoe2004a} could be thought of as material phases. Due to the closeness of the tasks, it is only reasonable to assume that some of the impact which deep learning has had on the field of computer vision will transfer to the field of topology optimization.

Led by this promise, recent years have seen a surge in publications on the application of deep learning \citep{lecun2015a} to topology optimization within a variety of problems such as minimum compliance \citep{hoyer2019a,zhang2019a,cang2019a,yu2019a,chandrasekhar2021a,nie2020a,kallioras2020a}, thermal compliance \citep{lin2018a,li2019a,li2020a}, micro-structure design \citep{abueidda2019a,tan2020a,wang2020a,kollmann2020a,chen2020a} and generative design \cite{oh2019a}. Despite these efforts, deep learning has yet to make a major impact within topology optimization, and only performs at a similar level to traditional topology optimization methods at very low design resolutions, or for highly restricted problem domains and boundary conditions, where the cost of generating a synthetic dataset and training a neural network is not prohibitive. Put bluntly, and due to the above reasons, none of the so far published deep learning end-to-end methods allow for a level of generalization which makes it advisable to use them for practical applications. Moreover, for end-to-end machine learning to have a significant impact on the field of topology optimization, it is clearly a requirement that it is not only faster and scales better, but also that the solutions produced are of comparable quality to those computed using more conventional optimization methods \citep{bendsoe2004a}. Furthermore, this must be true even when the network is provided inputs that are very different from those in the training data. This appears to be a very challenging problem, but attractive alternatives to end-to-end learning present themselves. For example, it is generally much easier to incorporate constraints and ensure a physically sound design in the classical mathematical programming methods, where only one design problem is considered at a time, whereas convolutional neural networks have shown state-of-the-art performance on tasks such as image denoising \cite{ulyanov2020a} and upsampling \cite{ledig2017a} where local information plays an important role. Hence, a combined method, in which the two approaches complement each other, seems to be a much more viable direction to pursue \citep{chi2021a}.\\

\begin{figure}[tb]
	\centering
	\includegraphics[width=\textwidth]{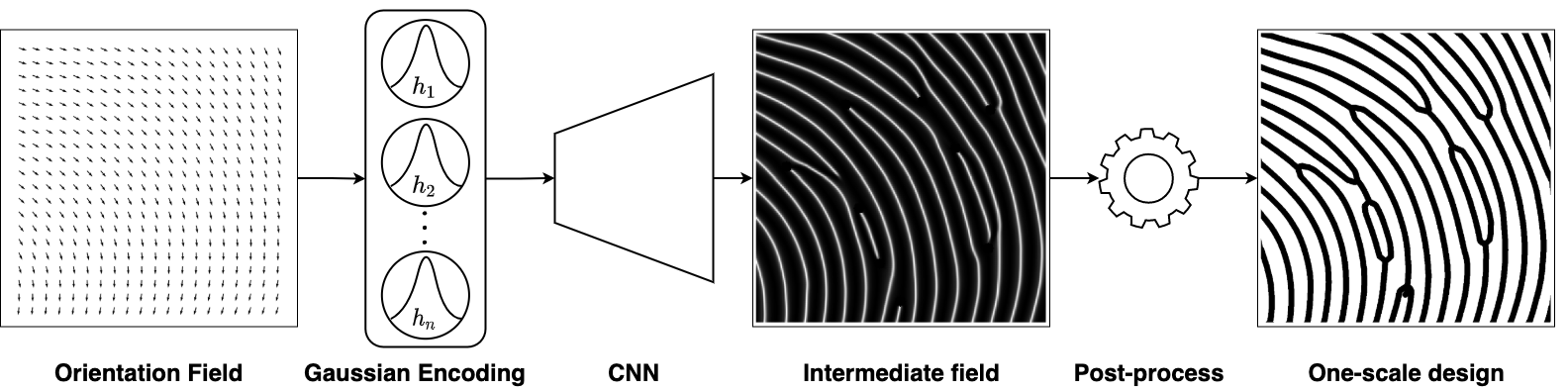}
	\caption{Data pipeline from orientation field to intermediate density field on to one-scale design.}
	\label{fig:data_pipeline}
\end{figure}

Here we report on a deep learning approach that allows for problem definition generalization and which is capable of producing high resolution and detailed designs. The key to this achievement lies in the fact that the proposed method does not provide an end-to-end topology optimization approach. Instead, it provides an alternative to traditional de-homogenization approaches \cite{pantz2008a,groen2018a,allaire2019a,groen2020a} by using the orientations from a homogenization-based topology optimization solution \citep{bendsoe1988a,groen2018a} as inputs for a fully convolutional neural network \cite{shelhamer2017a}, which upsamples them to an intermediate field $\tilde{\rho}$. The network is trained using a synthetic dataset generated by sampling the spatial gradient orientations from a field of low-frequency sines, and an encoding based on splitting the angular information into multiple channels is used to ease learning for the network. A custom loss function ensures that the intermediate field is periodic according to a predefined wave-length and that its spatial gradients are orthogonal to the input directions. Furthermore, two regularization terms are included that favor a solution with sufficient spatial variation, and where branching mainly takes place in the solid phase. Figure \ref{fig:data_pipeline} shows the data pipeline from the orientation field to the intermediate field and subsequent post-processing. The post-processing consists of a series of fast steps, all running in linear time, which project the desired lamination widths on the intermediate field to yield the de-homogenized one-scale design.\\
We remark, that solving the homogenization-based topology optimization problem is already very fast, i.e. the problem is close to convex and much information can be extracted, even from coarse finite element meshes \cite{groen2020a}. The current bottleneck in de-homogenization methods is the expensive solving of one large-scale least square problem per lamination direction. In this work, we propose a learning-based solution to this problem that promised significant time savings. With computation times around 2 seconds for projection onto a $2400\times1200$ mesh on a standard laptop and lots of potential for further improvements and porting to GPUs, this approach may pave the way for future real-time de-homogenization in the iterative homogenization-based optimization loop \cite{NobAagSig15}. \\

The structure of the paper is as follows; in Section \ref{sec:methods} the relevant methods are outlined including homogenization-based topology optimization, network architecture, loss function, and post-processing scheme. Section \ref{sec:data+input} presents the synthetic dataset generation and input encoding. In Section \ref{sec:results} results for several different minimum compliance examples are presented and compared to a reference solution provided by the homogenization-based method. Finally, a discussion of the strengths and weaknesses of the proposed method and potential improvements are discussed in Section \ref{sec:conclusion}.

\section{Methods} \label{sec:methods}

This section provides a brief introduction to homogenization-based topology optimization, followed by a detailed description of the proposed neural network and lamination width projection scheme.

\subsection{Homogenization-based topology optimization}

The seminal papers by Bends{\o}e and Kikuchi \cite{bendsoe1988a,Ben89} introduced the so-called homogenization-based topology optimization approach. Crucially, this method overcame the ill-posedness of element or node-based design parameterizations by introducing optimal, multi-scale, so-called rank-N   (or slightly sub-optimal rectangular hole) microstructures, oriented along principal stress directions. This approach was considered quite complex and also resulted in "grey" solutions that could not be manufactured. Therefore, and specifically after the simpler density-approach was proven physically admissible \cite{BenSig99}, homogenization-based topology optimization was largely abandoned in favor of simpler approaches. Now, where simpler density approaches have reached giga-scale resolution \cite{BaaSigAag20} that require access to supercomputers, and where advances in additive manufacturing techniques allow realization of efficient multi-scale structures, there has been a renewed interest in homogenization-based topology optimization approaches, that can be run efficiently on simple computers and who's output can be realized through novel de-homogenization approaches \cite{pantz2008a,groen2018a,allaire2019a,groen2020a}.

\begin{figure}[H]
	\centering
	\includegraphics[width=0.8\textwidth]{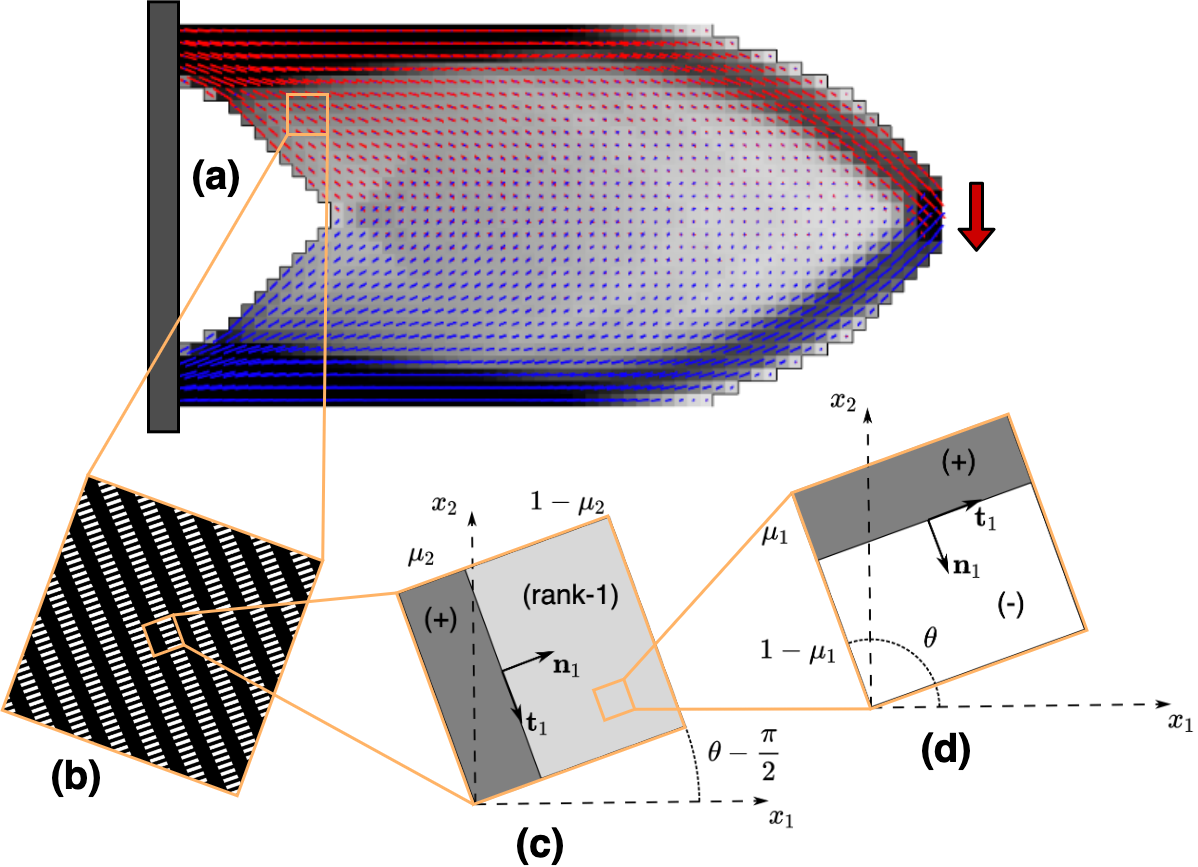}
	\caption{Parameterization of rank-2 microstructures in homogenization-based topology optimization. (a) Cantilever beam  with boundary conditions. (b) Rank-2 laminate structure. (c) Parameterization of rank-2 microstructure. (d) Parameterization of rank-1 microstructure made from a stiff material (+) and a weak material (-). }
	\label{fig:rank2}
\end{figure}

The homogenization-based approach to topology optimization can be explained through the visualization in Figure \ref{fig:rank2}. A design domain (a) is discretized into a relatively low number of elements and each element has three design variables. Lamination parameters $\mu_1$ and $\mu_2$ denote fractions of solid material at two different length scales and $\theta$ determines the orientation. For this so-called rank-2 material, homogenized effective properties can be determined in closed form (see e.g. \cite{bendsoe2004a}). Based on analytical gradient computations, these three design fields can be optimized, resulting in e.g. the cantilever beam structure in Figure \ref{fig:rank2}a, where grey color indicates optimized element-wise volume fractions and blue and red lines indicate direction and proportions of the two local lamination thicknesses. The exact formulation of the homogenization-based topology optimization approach used here follows next.

\subsubsection{Optimization problem}

The goal of the considered homogenization-based topology optimization problem is to minimize the structural compliance $\mathcal{C}$, i.e. the work done by external forces, subject to a volume constraint. This objective is augmented by two additional terms; an angle regularization term $\mathcal{F}_{\theta}$ \cite{groen2020a} and a penalization of the solid area \cite{giele2021a}. The design variables are the two element-wise lamination parameters $\pmb{\mu}_1$, $\pmb{\mu}_2$, the angle $\pmb{\theta}$, as well as an auxiliary material indicator field $\mathbf{s}$.\\

The full optimization problem, including constraints, is stated as
\begin{equation}
	\begin{aligned}
		\min\limits_{\pmb{\mu}_1,\pmb{\mu}_2,\pmb{\theta},\mathbf{s}}:&  \quad \frac{\mathcal{C}(\pmb{\mu}_1,\pmb{\mu}_2,\pmb{\theta},\mathbf{u}) + \gamma_{\theta}\mathcal{F}_{\theta}(\pmb{\theta})}{\mathcal{C}_{ref}} +  \Gamma \frac{V_{\Omega,\text{struct}}^e(\mathbf{s})}{V_{\Omega}}\\
		\text{s.t}:& \quad \mathbf{K}(\pmb{\mu}_1,\pmb{\mu}_2,\pmb{\theta}) \mathbf{u} = \mathbf{f} \\
		:& \quad \mathbf{v}^T \pmb{\rho}^d(\pmb{\mu}_1,\pmb{\mu}_2) - V_{max} \leq 0 \\
		:& \quad \mu_{min} \leq \pmb{\mu}_1,  \pmb{\mu}_2 \leq 1 \\
		:& \quad -4\pi \leq \pmb{\theta} \leq 4\pi
	\end{aligned}
	\label{eq:opt_problem}
\end{equation}
Here $\mathbf{K}$ is the stiffness matrix, $\mathbf{f}$ is the force vector, $\mathbf{u}$ is the solution to the equilibrium equations, $\mathbf{v}$ is a vector of element volumes, $\pmb{\rho}^d$ is a vector of element densities, $V_{max}$ is the maximum amount of allowed material and $\mathcal{C}_{ref}$ is a reference compliance. Finally, $\gamma_{\theta}$ and $\Gamma$ are constants used to scale the importance of each term in the objective and $\mu_{min}$ is the minimum bound on the admissible layer thickness, from here on referred to as the minimum relative thickness. \\

The relation between design variables and physical variables is
\begin{align}
	& \bar{\tilde{\pmb{\mu}}}_i^k = (\alpha_{out} + (1-\alpha_{out}) \tilde{\pmb{\mu}}_i)\bar{\tilde{\mathbf{s}}}^k \\
	& \pmb{\rho}^k =  \bar{\tilde{\pmb{\mu}}}_1^k  +  \bar{\tilde{\pmb{\mu}}}_2^k -  \bar{\tilde{\pmb{\mu}}}_1^k   \bar{\tilde{\pmb{\mu}}}_2^k \\
	& V_{\Omega,\text{struct}}^e(\mathbf{s}) = \int_{\Omega} \bar{\tilde{\mathbf{s}}}^e \mathrm{d}\Omega, 
\end{align}
where $\alpha_{out}=10^{-9}$ is a small value indicating the stiffness in the void phase, and $\Omega$ denotes the optimization domain. Here, $\tilde{\square}$ denotes filtered variables and $\bar{\tilde{\square}}$ denote projected variables. With this formulation, we ensure clear designs without non-physical lamination parameters in the interval $[0,\mu_{min}[$. For a detailed description of this approach, the readers are referred to the paper by Giele et al. \cite{giele2021a}.

\subsection{Convolutional neural network} \label{sec:conv_net}

The homogenization-based topology optimization solution contains a set of lamination parameters describing the underlying microstructure in each element, but does not in itself provide a physically realizable solution, as infinite periodicity of the microstructure is assumed within each element (Figure \ref{fig:rank2}b). Together with the element-wise varying lamination angles, this poses the difficult challenge of seamlessly connecting the microstructure in adjacent elements to form a mechanically sound and efficient de-homogenized design at finite periodicity. In the context of previous de-homogenization approaches, this is where the current bottleneck lies \citep{pantz2008a,groen2018a,allaire2019a}. The most expensive part of existing de-homogenization approaches is the solution of a least-squares problem on an intermediate mesh to identify two auxiliary fields with gradients corresponding to the normals of the two lamination directions. Hence, it is interesting to investigate if this task can be performed more efficiently with a learning-based approach. \\

The most straightforward way to utilize a neural network for solving this task would be to do it in a supervised manner \citep{norvig2010a}. In the supervised learning case, the network takes the lamination parameters as input and outputs a de-homogenized design which is then evaluated against a ground-truth de-homogenized design generated using existing de-homogenization methods \citep{groen2018a,allaire2019a}, i.e. the neural network $G$ creates a mapping
\begin{equation}
	\pmb{\rho} = G(\pmb{\mu}_1,\pmb{\mu}_2,\pmb{\theta})
\end{equation}
and is evaluated using an error estimate, such as the mean-squared error, between the predicted and ground-truth design
\begin{equation}
	\mathcal{L}_{MSE} = \frac{1}{n} \sum_{i=1}^{n}(\pmb{\rho}-\pmb{\rho}_{gt})^2
	\label{eq:mse_loss}
\end{equation}
Where $\pmb{\rho}_{gt}$ denotes the ground-truth design.\\

In practice, there are several downsides to such an approach. First of all, a very large amount of computation is required in order to generate a dataset of ground-truth designs with sufficient resolution and quality. Secondly, even though de-homogenization methods have improved tremendously in recent years most methods still require some degree of manual intervention, which is both cumbersome and undesirable to do for thousands of samples. Also, an entirely target-based loss function only ensures pixel-wise accuracy, but does not ensure the structural integrity of the design as a whole. Thus, if a few crucial pixels are predicted with low accuracy the design may contain structural disconnects, but exhibit a low geometrical error overall. Of these two issues, the dataset generation is the most pressing, as minor structural disconnects can often be fixed by a subsequent post-processing scheme. 
Therefore, and due to the large amount of computation and manual work needed to make a supervised approach work, an unsupervised approach \cite{ghahramani2004a} is taken in this paper. Our solution is based on the observation that the mechanical soundness of the structure is provided by the homogenization-based solution. Thus, our convolutional network needs only to produce structures of a given periodicity from the lamination orientations provided by the homogenization-based solution, whereas the lamination widths are projected onto the design in a post-processing step. This problem greatly resembles texture synthesis which crucially allows us to train the network using synthetic fields. However, initial tests using the orientation field directly as input for the neural network did not show promising results, and thus an alternative encoding based on splitting the angular field into multiple channels each modeled by a Gaussian was developed as detailed in Section \ref{sec:data+input}.\\

Given the activations in each channel the network $G$ performs the following mapping
\begin{equation}
	\tilde{\pmb{\rho}} = G(\mathbf{H})
\end{equation}
Where $\mathbf{H}$ are the Gaussian activations, and $\tilde{\pmb{\rho}}$ is an intermediate field  who's gradients follow lamination orientations, i.e. $|\nabla \tilde{\rho}  \cdot \vec{\mathbf{e}}_{\theta}| \approx 0$. Note that for a rank-2 microstructure description the lamination orientations correspond to the principal stress orientations.\\

A fully convolutional neural network \cite{shelhamer2017a} is chosen as the desired network architecture as it allows an arbitrary input size, and requires much fewer learning parameters compared to a fully connected neural network. In the bottom layer of the network, a series of Residual Neural Network (ResNet) blocks \cite{he2015a} are used to extract local features from the input orientations. To accurately represent the orientation information in the intermediate field a total upsampling factor of $\times8$ is performed. Each upsampling layer consists of a nearest-neighbour upsampling followed by two convolutional layers, which have proven less prone to checkerboard-artifacts compared to other upsampling methods such as pixel-shuffle or transposed convolution \cite{mordvintsev2015a}. The final output layer consists of a $3\times3$ kernel followed by a Sigmoid activation to scale the output between 0 and 1. More information on the exact network architecture can be found in \ref{app:net_arch}.\\

\begin{figure}[tb]
     \centering
     \begin{subfigure}[b]{0.35\textwidth}
         \centering
         \includegraphics[width=\textwidth]{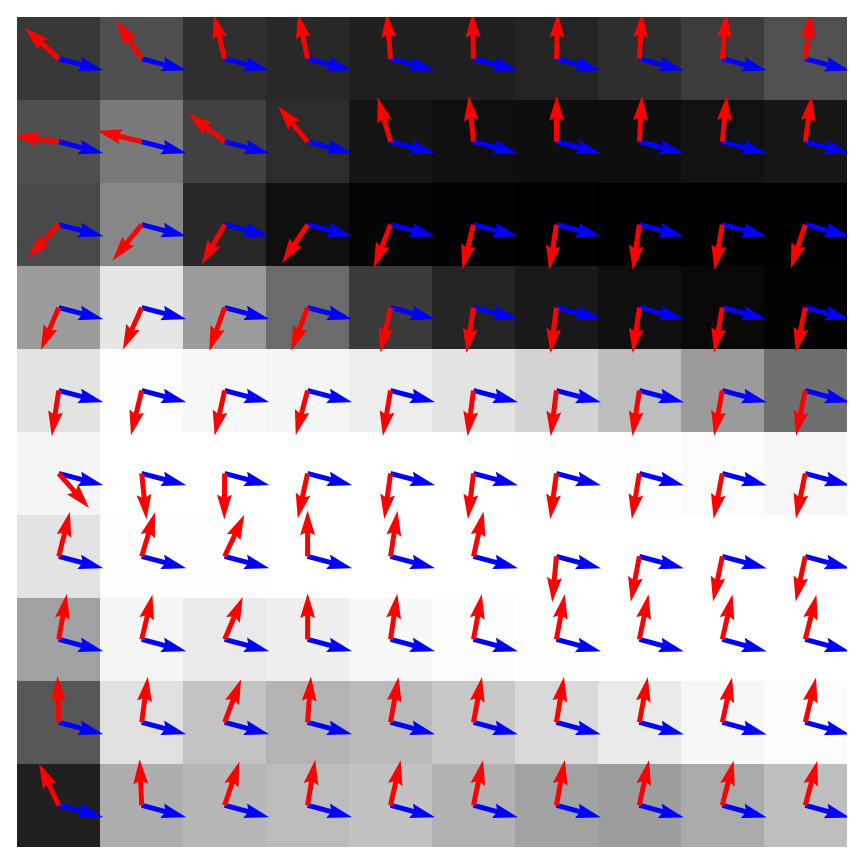}
         \caption{$\mathbf{e}_{\theta}$ (\textcolor{blue}{blue}), $\mathbf{e}_{\tilde{\rho}}$ (\textcolor{red}{red}) plotted on top of $\tilde{\pmb{\rho}}$.} 
         \label{fig:rho_tilde_grad_vecs}
     \end{subfigure}
     \qquad
     \begin{subfigure}[b]{0.37\textwidth}
         \centering
         \includegraphics[width=\textwidth]{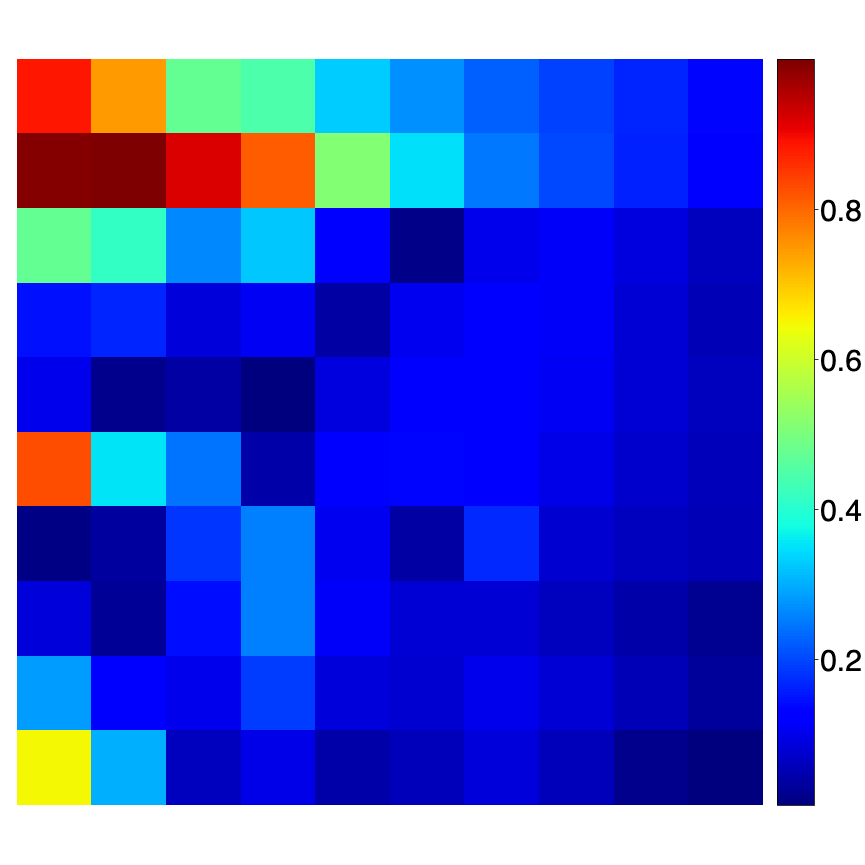}
         \caption{Element-wise $\mathcal{L}_{dot}$}
         \label{fig:elem_dot_product_loss}
     \end{subfigure}
    \caption{Visual representation of the dot-product loss near the edge of the domain when the network has been trained.}
    \label{fig:dot_product_loss}
\end{figure}

As a loss function, the dot-product between the orientations from the homogenization-based solution and the normalized gradients of the intermediate field is used
\begin{equation}
	\mathcal{L}_{dot} =  \hat{\vec{\mathbf{e}}}_{\tilde{\rho}}(\mathbf{x}) \cdot \vec{\mathbf{e}}_{\theta}(\mathbf{x})
	\label{eq:dot_loss}
\end{equation}
Where $\mathbf{e}_{\theta}$ is the vectorized version of the lamination angle, and $\hat{\mathbf{e}}_{\tilde{\rho}}=\nabla \tilde{\pmb{\rho}}/ \lVert \nabla \tilde{\pmb{\rho}} \rVert_2$ is calculated using the Sobel operator. To avoid very small gradient magnitudes from influencing numerical stability during training a mean filter is applied to $\nabla \tilde{\pmb{\rho}}$ before normalization. Figure \ref{fig:dot_product_loss} shows a visual representation of the dot-product loss for a small patch of $\tilde{\pmb{\rho}}$ near the edge of the domain. It can be seen that generally, the loss is higher near the edge as the Sobel operator does not provide an accurate estimate of the gradient here, while orientations further from the domain boundaries are near orthogonal. To counteract this issue, replication padding is added to the neural network input.\\

To control the periodicity of the output field a windowed Fourier transformation
\begin{equation}
	\tilde{f}(x,y) = \mathcal{F}(\tilde{\rho}(x,y) \cdot w(x,y) )
\end{equation}
is utilized as the basis for an additional term in the loss function. Here a Hamming window $w$ has been used to account for spectral leakage \cite{harris1978a}. This additional term measures the energy inside a specified frequency band  compared to the total energy of the spectrum
\begin{equation}
	E_{\omega} = \frac{\sum_{i_{\omega} } \sum_{j_{\omega}} | \tilde{f}_{i_{\omega},j_{\omega}} |^2 }{\sum_i \sum_j | \tilde{f}_{i,j} |^2}
\end{equation}

In terms of de-homogenization it is most intuitive to think of this frequency band as describing the desired wave-length $\varepsilon$. Since an exact fulfillment of the desired wave-length leads to a very narrow frequency band, which is impractical during optimization, it is desirable to assign a small fixed width of $b$ pixels to the frequency band. Given a square image of width $H$ the frequency interval is simply
\begin{equation}
	\{i_{\omega},j_{\omega} \} = \left\{  \mathbb(i,j) \hspace{1mm} | \hspace{1mm}   \frac{H}{\varepsilon} - \frac{b}{2} 
	< \sqrt{i^2+j^2}< \frac{H}{\varepsilon} + \frac{b}{2} \right\}
\end{equation}
Where we assume the zero-frequency is at the origin. The spectral loss is included in the total loss function as 

\begin{equation}
	\mathcal{L}(\mathbf{e}_{\tilde{\rho}},\mathbf{e}_{\theta},\tilde{\mathbf{f}}) = \mathcal{L}_{dot}(\mathbf{e}_{\tilde{\rho}},\mathbf{e}_{\theta}) 
	- \lambda_{\omega} E_{\omega}(\tilde{\mathbf{f}}) 
	\label{eq:ldot+lomega}
\end{equation}
Here  $\lambda_{\omega}$ is a weight factor and a minus is used in front of $E_{\omega}$ as we seek to maximize the energy inside the specified frequency band, but we seek to minimize the total objective.\\

If no regularization is used, the objective in eq. \ref{eq:ldot+lomega} tends to favor solutions with very low spatial variation in terms of magnitude. This might lead to numerical instabilities when performing normalization of the image gradient orientations, especially since the computations are performed on a single-precision GPU. Two different approaches have been tested in order to counteract an overly smooth representation of $\tilde{\pmb{\rho}}$. The simplest approach is to add weak Gaussian noise to the input orientation fields, before performing the input encoding. This helps the network learn more robust representations of the intermediate field, essentially pushing the values towards the extremes of the Sigmoid function, where small variations have less impact. In practice the Gaussian noise alone is often enough to ensure sufficient spatial variation. In addition to the Gaussian noise a total variation term
\begin{equation}
	\mathcal{V}(y) = \sum_{i,j} (y_{i+1,j} - y_{i,j})^2 + (y_{i,j+1} - y_{i,j})^2 
	\label{eq:TV}
\end{equation}
is added to the loss to enforce a predefined total variation of the intermediate field.\\

It is important to note that the total variation term on its own does not provide an explicit way of controlling the periodicity of the intermediate field, as the total variation may be increased in multiple ways, i.e. either through increasing the constrast, the periodicity or a mixture of both. Thus, the total variation and spectral losses should be used in conjunction
\begin{equation}
	\mathcal{L}(\tilde{\mathbf{\rho}},\mathbf{e}_{\tilde{\rho}},\mathbf{e}_{\theta},\tilde{\mathbf{f}}) = \mathcal{L}_{dot}(\mathbf{e}_{\tilde{\rho}},\mathbf{e}_{\theta}) 
	- \lambda_{\omega} E_{\omega}(\tilde{\mathbf{f}})
	+ \lambda_{\tau} \left\lVert \frac{\mathcal{V}(\tilde{\pmb{\rho}})}{\tau} - 1 \right\lVert_2
	\label{eq:ldot+lomega+ltv}
\end{equation}
Where $\tau$ is the desired total variation, and $\lambda_{\tau}$ is a weight factor.\\

\begin{figure}[tb]
	\centering
	\begin{subfigure}[b]{0.40\textwidth}
		\centering
		\includegraphics[width=\textwidth]{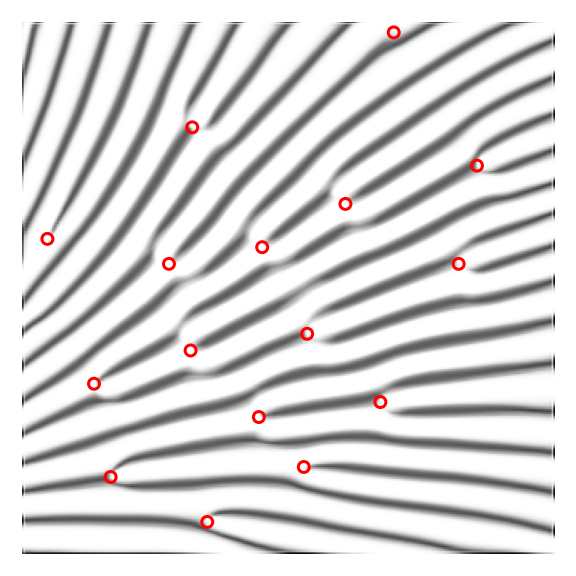}
		\caption{$\mathcal{L}_b$ maxima plotted on top of $\tilde{\pmb{\rho}}$}
		\label{fig:Idot_maxima}
	\end{subfigure}
	\qquad
	\begin{subfigure}[b]{0.45\textwidth}
		\centering
		\includegraphics[width=\textwidth]{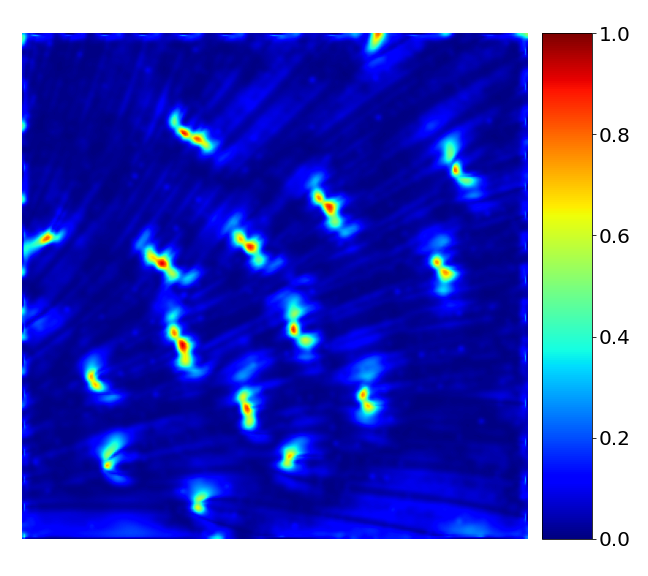}
		\caption{$\mathcal{L}_b$ loss surface} 
		\label{fig:Idot_indicator}
	\end{subfigure}
	\caption{An intermediate field $\tilde{\pmb{\rho}}$ and the corresponding branching loss surface $\mathcal{L}_b$. Red circles indicate maxima of the $\mathcal{L}_b$ loss surface.}
	\label{fig:branching_loss}
\end{figure}

For orientational changes to take place in the intermediate field, the periodicity must be increased or branches occur. Since the periodicity is fixed by the spectral loss, branching is the only option. Without any regularization, areas around a branching point often take intermediate values. This is undesirable if a thresholding approach is later used to obtain a fully solid-void design as it might lead to disconnects. Thus, it is preferable to drive branches towards the solid phase. One possible way to do this is to use the dot-product loss in eq. \ref{eq:dot_loss} as an indicator function for branching. This is based on the observation that peaks in the dot-product loss most often coincide with branching, see Figure \ref{fig:branching_loss}. 

Using the dot-product loss surface directly as an indicator function introduces an unnecessary amount of noise to the optimization, as there are lots of small peaks in the loss-surface not related to branching. Furthermore, the error will often be large near the edge of the domain, as the Sobel operator provides a poor estimate of the gradients in this area. To create a cleaner indicator function, the loss surface is first smoothed using a small Gaussian kernel $g$
\begin{equation}
    \mathbf{I}_{dot}(x,y) = \mathcal{L}_{dot}(x,y)*g(x,y)
\end{equation}

Subsequently, all values in $\mathbf{I}_{dot}$ closer than 5 pixels to the edge of the domain are set to zero. Now, a loss term favoring branching in the solid phase may be formulated based on the indicator function 

\begin{equation}
    \mathcal{L}_{b}(\mathbf{I}_{dot},\tilde{\pmb{\rho}}) = \mathbf{I}_{dot} \odot (1-\tilde{\pmb{\rho}})
    \label{eq:branching_loss}
\end{equation}
Where $\odot$ indicates element-wise multiplication, and $1-\tilde{\pmb{\rho}}$ is used such that only dot-product errors in the void-phase are penalized. This naturally drives the intermediate field towards values very close to one, but still far enough apart to provide meaningful image gradients.\\

Including the branching loss in eq. \ref{eq:ldot+lomega+ltv}  leads to the final loss function used for this work, i.e.
\begin{equation}
	\mathcal{L}(\tilde{\pmb{\rho}},\mathbf{e}_{\tilde{\rho}},\mathbf{e}_{\theta}, \tilde{\mathbf{f}}, \mathbf{I}_{dot}) = \mathcal{L}_{dot}(\mathbf{e}_{\tilde{\rho}},\mathbf{e}_{\theta}) 
	- \lambda_{\omega} E_{\omega}(\tilde{\mathbf{f}})
	+ \lambda_{\tau} \left\lVert \frac{\mathcal{V}(\tilde{\pmb{\rho}})}{\tau} - 1 \right\lVert_2 
	+ \lambda_{b} \mathcal{L}_{b}(\mathbf{I}_{dot},\tilde{\pmb{\rho}})
	\label{eq:ldot+lomega+ltv+lb}
\end{equation}
Where $\lambda_{\omega}$, $\lambda_{\tau}$ and $\lambda_b$ are scaling parameters, used to weigh the importance of the spectral, total variation and branching loss in the full objective.\\

However, minimizing all four terms in eq. \ref{eq:ldot+lomega+ltv+lb} simultaneously often leads to poor local minima, and requires careful balancing of the four objectives to obtain useful results. To remedy this issue the training is performed in two steps; in the first step the network is trained with only the orientation, spectral and total variation losses enabled, i.e. $\lambda_b=0$. This ensures that the generated intermediate field follows the input orientations, and the prescribed periodicity. In the second step the spectral loss is disabled, and the branching loss is enabled instead, i.e. $\lambda_{\omega}=0$. The network is then initialized with the weights from the first training step and trained until convergence. \\
This procedure makes sense as the branching loss relies on branches being the main source of error in the orientation loss, which is not the case during early stages of the first training phase. On the other hand the spectral loss disturbs the generation of well-defined branches, as they violate the prescribed periodicity. The overall training procedure is outlined in Algorithm \ref{alg:two_step_train}.\\

\begin{algorithm}[H]
	\SetKwFunction{KaimingInit}{KaimingInit}
	\SetKwFunction{DataLoader}{DataLoader}
	\SetKwFunction{ImgGrad}{ImgGrad}
	\SetKwFunction{DFT}{DFT}
	\SetKwFunction{GaussKernel}{GaussKernel}
	\SetAlgoLined
	\emph{First step - orientation, spectral and total variation losses}\\
	$\mathbf{w}_1$ $\leftarrow$ \KaimingInit{} \\
	$\mathcal{L}_1(\tilde{\pmb{\rho}},\mathbf{e}_{\tilde{\rho}},\mathbf{e}_{\theta}, \tilde{\mathbf{f}}) = \mathcal{L}_{dot}(\mathbf{e}_{\tilde{\rho}},\mathbf{e}_{\theta}) 
	- \lambda_{\omega} E_{\omega}(\tilde{\mathbf{f}})
	+ \lambda_{\tau} \left\lVert \frac{\mathcal{V}(\tilde{\pmb{\rho}})}{\tau} - 1 \right\lVert_2$\\
	\For{$i \leftarrow 1$ \KwTo $n_{1,epochs}$}{
		$\mathbf{H}, \mathbf{e}_{\theta}$ $\leftarrow$ \DataLoader{$i$} \\
		$\tilde{\pmb{\rho}} \leftarrow G(\mathbf{H})$ \\ 
		$\mathbf{e}_{\tilde{\rho}}$ $\leftarrow$ \ImgGrad{$\tilde{\pmb{\rho}}$} \\ 
		$ \tilde{\mathbf{f}}$ $\leftarrow$ \DFT{$\tilde{\pmb{\rho}}$} \\
		$\mathbf{w}_1^{(i)} \leftarrow \mathbf{w}_1^{(i-1)} - \eta \nabla \mathcal{L}_1(\tilde{\pmb{\rho}},\mathbf{e}_{\tilde{\rho}},\mathbf{e}_{\theta}, \tilde{\mathbf{f}})$
	}
	\emph{Second step - orientation, total variation and branching losses}\\
	$\mathbf{w}_2 \leftarrow \mathbf{w}_1$ \\
	$\mathcal{L}_2(\tilde{\pmb{\rho}},\mathbf{e}_{\tilde{\rho}},\mathbf{e}_{\theta}, \mathbf{I}_{dot}) = \mathcal{L}_{dot}(\mathbf{e}_{\tilde{\rho}},\mathbf{e}_{\theta}) 
	+ \lambda_{\tau} \left\lVert \frac{\mathcal{V}(\tilde{\pmb{\rho}})}{\tau} - 1 \right\lVert_2 
	+ \lambda_{b} \mathcal{L}_{b}(\mathbf{I}_{dot},\tilde{\pmb{\rho}})$\\
		\For{$i \leftarrow 1$ \KwTo $n_{2,epochs}$}{
		$\mathbf{H}, \mathbf{e}_{\theta}$ $\leftarrow$ \DataLoader{$i$} \\
		$\tilde{\pmb{\rho}} \leftarrow G(\mathbf{H})$ \\ 
		$\mathbf{e}_{\tilde{\rho}}$ $\leftarrow$ \ImgGrad{$\tilde{\pmb{\rho}}$} \\ 
		$\mathbf{I}_{dot}$ $\leftarrow$ \GaussKernel{$\tilde{\pmb{\rho}}$}\\
		$\mathbf{w}_2^{(i)} \leftarrow \mathbf{w}_2^{(i-1)} - \eta \nabla \mathcal{L}_2(\tilde{\pmb{\rho}},\mathbf{e}_{\tilde{\rho}},\mathbf{e}_{\theta}, \mathbf{I}_{dot})$
	}
	\caption{Two-step training procedure}
	\label{alg:two_step_train}
\end{algorithm}

\subsection{Lamination width projection}

The homogenization-based topology optimization is based on optimally oriented rank-2 microstructures as described in Section 2.1, However, to achieve practically realizable single-scale microstructures, we convert the rank-2 microstructures to a single scale rectangular-hole microstructures with little loss in stiffness \cite{TraSigGro18} as discussed in \cite{groen2018a}. \\
 
Given the intermediate field $\tilde{\pmb{\rho}}_1$ for a single lamination direction a post-processing procedure is applied in order to project the desired lamination width onto the intermediate field. Below, the post-processing procedure for a single direction is outlined, but the procedure remains entirely the same for the other direction. \\
To make sure that the smallest features in $\tilde{\pmb{\rho}}_1$ are resolved by at least three pixels a bilinear upsampling to a finer mesh $\mathcal{T}^f$ is first performed. An estimate of the required upsampling factor needed to ensure this can be calculated as
\begin{equation}
	m_{up} = \left\lceil \frac{h_{min}}{\varepsilon_i \cdot \mu_{min}} \right\rceil
\end{equation}
Where $\mu_{min}$ is the minimum relative thickness, $h_{min}$ is the minimum feature size in pixels and $\varepsilon_i$ is the wave-length in pixels/period on the intermediate mesh $\mathcal{T}^i$.\\

\begin{figure}[H]
	\centering
	\includegraphics[width=\textwidth]{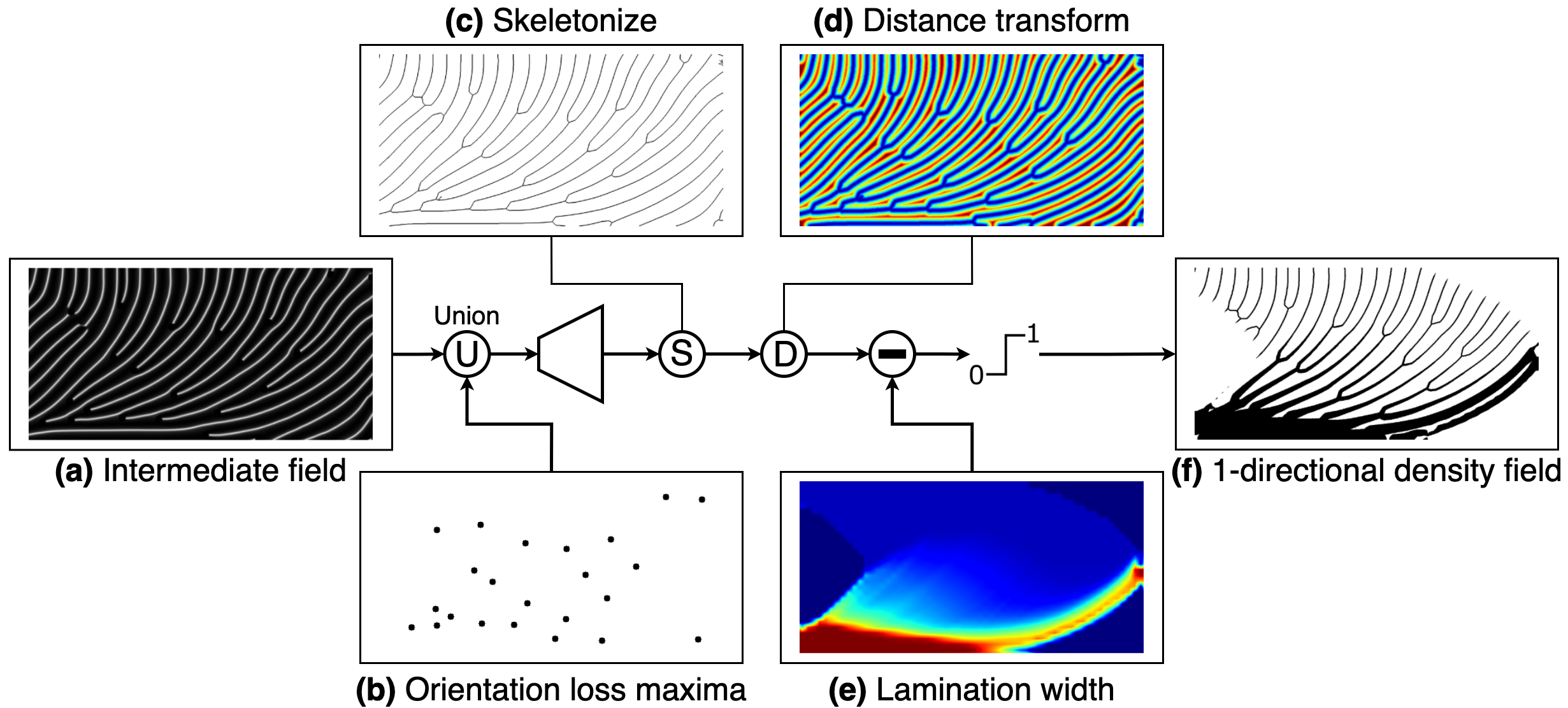}
	\caption{Graphical illustration of the post-processing procedure}
	\label{fig:postprocess_scheme}
\end{figure}

Once $\tilde{\pmb{\rho}}_1$ is of reasonable resolution the skeleton (c.f. Figure \ref{fig:postprocess_scheme}c) of the solid phase is extracted. However, in order to do so a few preparation steps are made. First, $\tilde{\pmb{\rho}}_1$ is normalized to the range 0-1 as the network output does not always utilize the entire range of the Sigmoid. Second, branching regions are solidified by extracting all maxima above a certain threshold in the dot-product loss surface, and setting all pixels in a radius of $\varepsilon_i/4$ to 1, c.f. Figure \ref{fig:postprocess_scheme}b. Lastly, the upsampled intermediate field is converted to binary image by setting all values above the mean to 1. Once the preparations steps have been performed, a skeletonization \cite{zhang1984a} is performed on the binary image to obtain the underlying structure of the intermediate field. Since skeletonization thins the solid part of the density field to one-pixel thickness, a dilation with a pixel-radius of 1 is applied to avoid point connections in diagonal bars. The dilated skeleton now serves as the minimum thickness for all members in the structure. \\

Having established the underlying skeleton and minimum thickness of all members in the structure a width is now assigned to each member. Here a distance transform $D_1$ (c.f. Figure \ref{fig:postprocess_scheme}d) is used to turn the skeleton image into a distance field. 
Given the distance field a thickness can be assigned to each member of the skeleton by using the lamination width $\mu_1$ to adaptively threshold the distance field
\begin{equation}
	\rho_1(\mathbf{x}) = \mathbb{H}\left(\mu_1(\mathbf{x}) - D_1(\mathbf{x})\right)
\end{equation}
Where $\mathbb{H}(\mathbf{x})$ is the Heaviside step function.\\

There are a few caveats to this procedure. First off, due to branching the periodicity in $\tilde{\pmb{\rho}}_1$ is not entirely uniform. This means that if $D_1$ is normalized to the 0-1 range outliers will influence the normalization, and cause an erroneous projection. To remedy this issue a histogram on $D_1$ can be used to identify outlier values, and determine an appropriate clipping value before normalizing the distance field. Secondly, $\mu_1$ should be clipped according to $\frac{h_{min}}{P \cdot m_{up}}$ and normalized subsequently, such that the minimum feature size corresponds to the minimum width resolvable by the mesh. Figure \ref{fig:postprocess_scheme} gives an overview of the entire post-processing procedure from the intermediate field to the 1-directional density field.\\ 

Once the density fields for both lamination orientations have been calculated the global density field can be obtained as the union between the two fields
\begin{equation}
    \rho(\mathbf{x}) = \min \left\{\rho_1(\mathbf{x})+\rho_2(\mathbf{x}),1 \right\}
\end{equation}

\section{Dataset and input encoding} \label{sec:data+input}

One of the absolute strengths of the method proposed in this paper, is that the training data does not rely on the physics or the underlying structural optimization problem. When training a neural network, using data from the desired application domain, in most cases, leads to the best performance. In our case this type of data would correspond to a set of orientation fields generated using the homogenization-based approach. However, such orientation fields are both expensive to obtain, and might contain singularities, especially around regions where load and boundary conditions have been applied. These singularties often result in non-integrable vector fields, which are not suited for de-homogenization \cite{stutz2020a}. To avoid singularities and lessen the computational burden a synthethic dataset is used instead.\\

To obtain a synthetic orientation field, we first create a synthetic scalar field, $F$,  by summing products of low frequency sine functions and obtain the needed vector field by taking the gradient:

\begin{align}
	F(x,y) &= \sum_n \sum_m c_{n,m} \sin \left(\frac{n \pi x}{x_L} \right) \sin \left(\frac{m \pi y}{y_L} \right) \label{eq:fourier_expansion} \\
	v_x(x,y) &= \frac{\partial F(x,y)}{\partial x} \label{eq:dFdx} \\
	v_y(x,y) &= \frac{\partial F(x,y)}{\partial y} \label{eq:dFdy}
\end{align}

Since the gradient magnitude is of no importance to the orientation, the gradients are normalized to yield

\begin{equation}
	\mathbf{e}(\mathbf{x}) = \mathbf{v}(\mathbf{x}) / \lVert \mathbf{v}(\mathbf{x)} \rVert_2
	\label{eq:e_norm}
\end{equation} 

\begin{figure}[tb]
	\centering
	\begin{subfigure}[b]{0.4\textwidth}
		\centering
		\includegraphics[width=\textwidth]{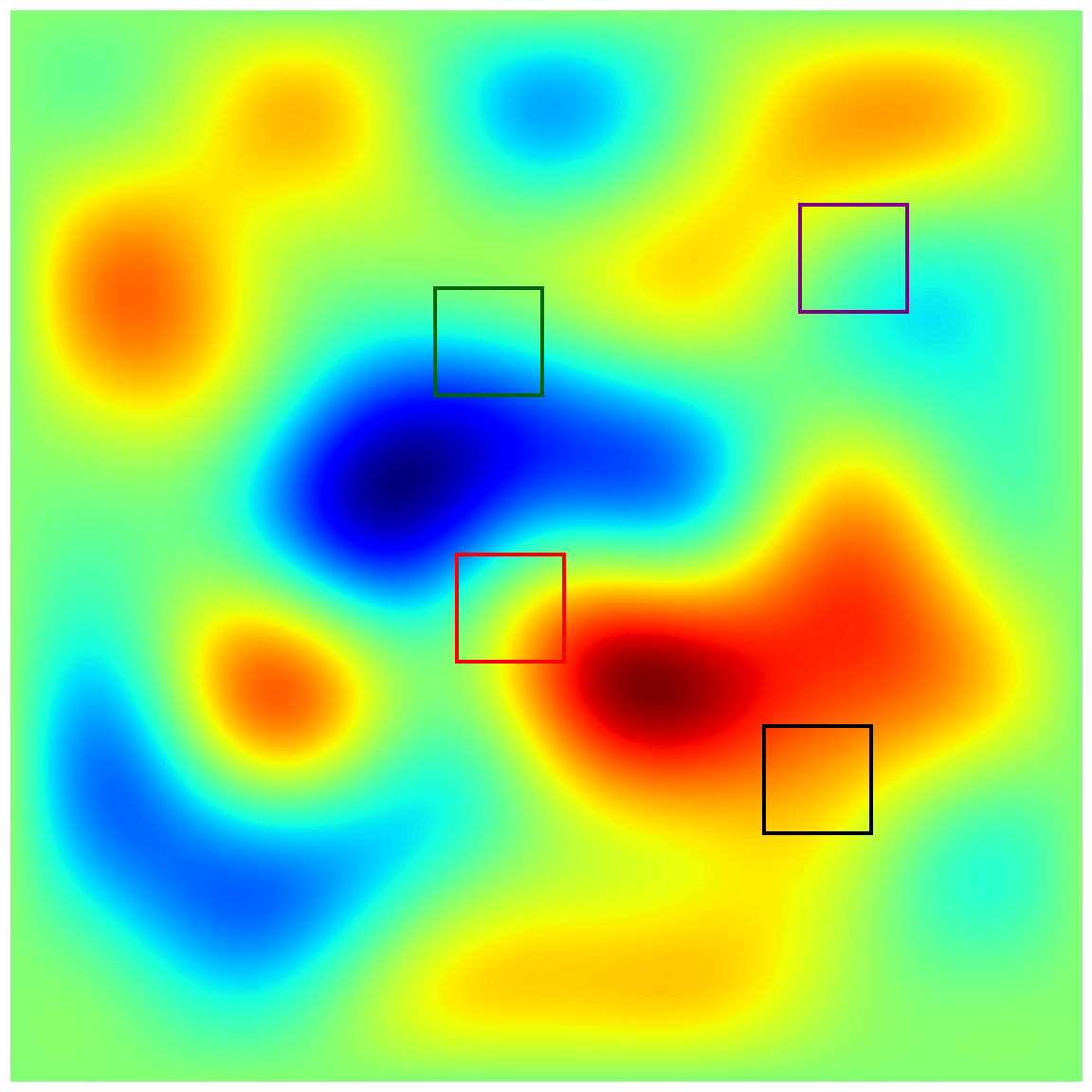}
		\caption{Global field}
		\label{fig:global_fourier_field}
	\end{subfigure}
	\qquad
	\begin{subfigure}[b]{0.4\textwidth}
		\centering
		\includegraphics[width=\textwidth]{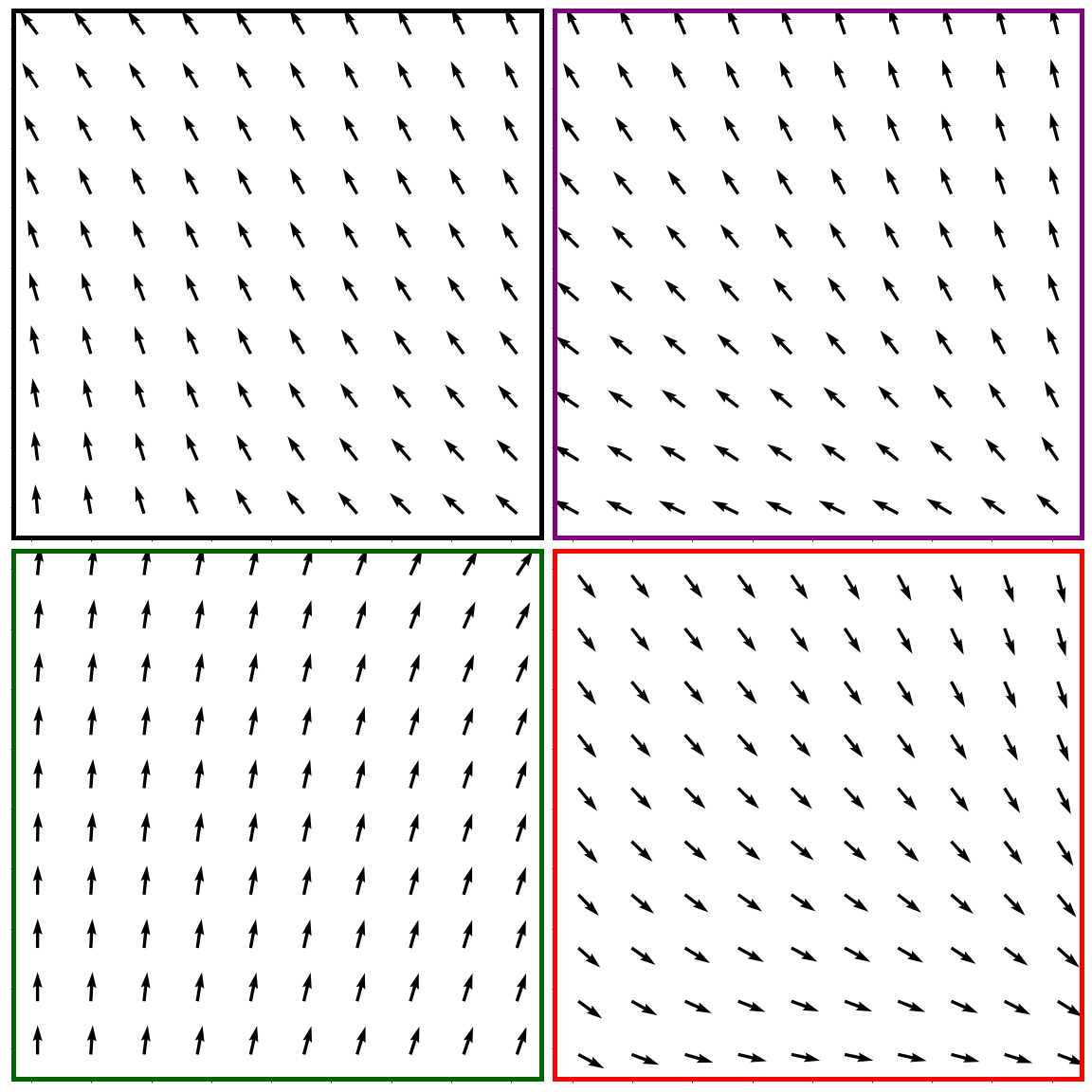}
		\caption{Patch orientation vectors}
		\label{fig:patch_grad_vecs}
	\end{subfigure}
	\caption{The global field composed of low-frequency sines and four randomly sampled orientation vector patches. Notice that orientation vectors have been subsampled to improve readability of the plot.}
	\label{fig:fourier_field_sampling}
\end{figure}

From the global orientation field generated using eqs. (\ref{eq:fourier_expansion}-\ref{eq:e_norm}) smaller patches with different aspect ratios may be sampled. To ensure a somewhat smooth orientation field within each patch a constraint on the maximum angular change $\theta_{max}$ between neighboring orientations is introduced. This constraint is enforced by resampling a patch if the constraint is violated, and also ensures that the singularities around local extrema in the global field are avoided. Furthermore, the global orientation field is resampled every $N_{rs}$ iterations with a random number of sine contributions from the set $C=\{6,8,10\}$. Figure \ref{fig:fourier_field_sampling} shows an example of four patches sampled using the above mentioned approach, while Table \ref{tab:dataset_params} provides an overview of the parameters used for the sampling procedure.\\

\begin{table}[H]
	\centering
	\begin{tabular}{c|c|c|c|c}
		Global field size & Patch sizes & $C$ & $\theta_{max}$ & $N_{rs}$ \\ \hline
		H=W=800 & \makecell{80x80\\ 60x120 \\ 40x160} & \makecell{$n=m=6$\\ $n=m=8$ \\ $n=m=10$} & 25 deg & 100 \\ \hline
	\end{tabular}
	\caption{Parameters used to sample the synthetic dataset.}
	\label{tab:dataset_params}
\end{table}

From Figure \ref{fig:patch_grad_vecs} it can be seen that the variation within each patch is generally not that large. Thus, to reduce computation and memory consumption the orientation vectors are subsampled to half the resolution using a 2x2 block average. Notice, even though the orientation vectors are subsampled it remains important to use a relatively large patch size to capture sufficient variation in the global field.\\

Initial tests using the orientation field $\mathbf{e}$ directly as input for the neural network did not show promising results, and thus an alternative encoding based on splitting the angular input into $N$ different channels was used. Each channel is now modeled using a Gaussian
\begin{equation}
	h = \exp \left( -\frac{(x-c)^2}{r^2} \right)
\end{equation}
with centers $c$ equally spaced in the interval $[0;\pi]$ and a kernel radius of $r=2\pi/N$. For this particular case $x$ indicates the input angle. \\

\begin{figure}[tb]
     \centering
     \begin{subfigure}[b]{0.4\textwidth}
         \centering
         \includegraphics[width=\textwidth]{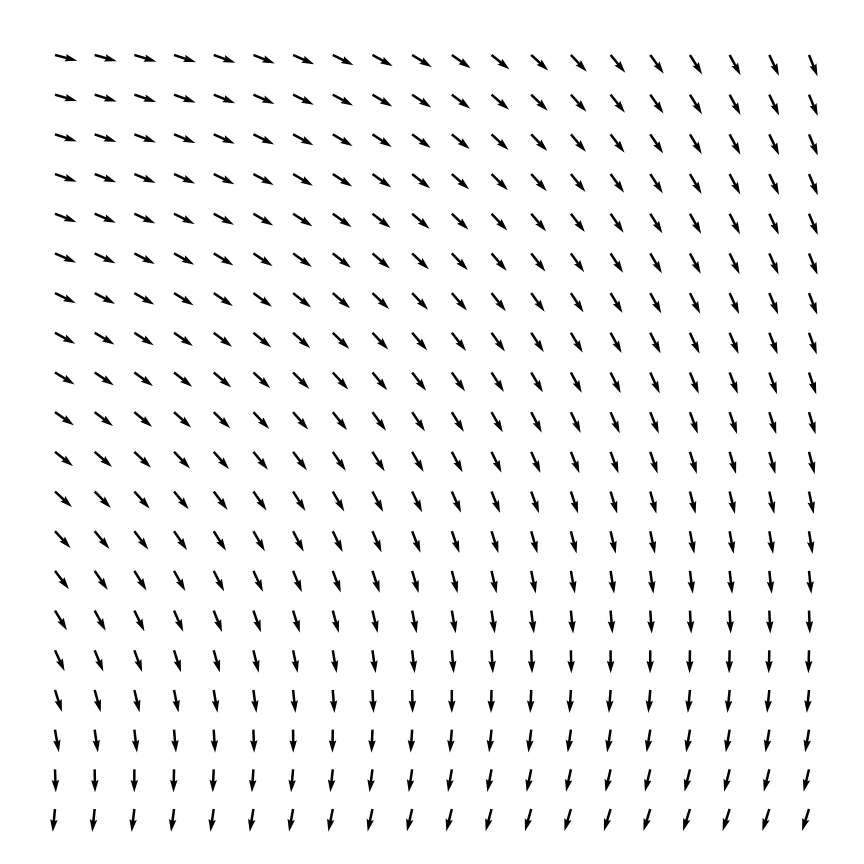}
         \caption{Orientation field}
         \label{fig:orientation_field}
     \end{subfigure}
     \hfill
     \begin{subfigure}[b]{0.58\textwidth}
         \centering
         \includegraphics[width=\textwidth]{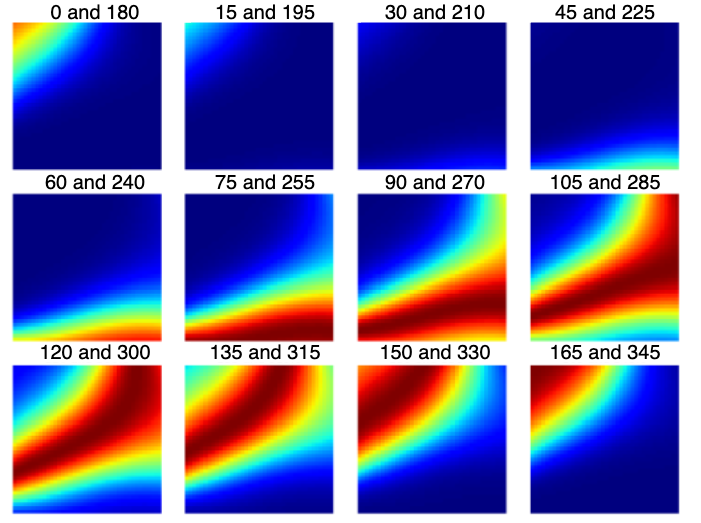}
         \caption{Spatial activations}
         \label{fig:rbf_activations}
     \end{subfigure}
    \caption{An orientation field and the corresponding spatial activations in each of the input encoding channels. }
    \label{fig:input_encoding}
\end{figure}

The encoding constitutes a mapping $\mathbb{R}^{1} \rightarrow \mathbb{R}^{N}$ from the angular information of $\mathbf{e}$ to an activation in each of the $N$ channels and is applied in an element-wise manner. This type of input encoding bears resemblance to binning, but ensures a smooth representation of the activation in each channel as the contribution of the angular input in each channel is weighted by the Gaussian. Figure \ref{fig:input_encoding} shows the activations in each of the channels for $N=12$ and an orientation field with a close to $90^{\circ}$ phase shift.\\ 
There are a few caveats which must be taken into account when performing this type of encoding. First, $\mathbf{e}$ is a 2-directional field, and thus invariant to rotations of $\pi$. This means that any angular values outside the $[0;\pi]$ range, in which the Gaussians are defined, can simply be shifted to the equivalent value inside the range. Second, to preserve the $\pi$-periodicity any activation in one end of the interval must be properly reflected in the other end of the interval. This is done by introducing a set of support Gaussians which extend the $[0;\pi]$ interval by $3r$ in each end. Any activations in these support functions are mapped to the appropriate channel in the opposite end of the interval. 

\section{Numerical examples} \label{sec:results}

\begin{figure}[tb]
	\centering
	\begin{subfigure}[t]{0.43\textwidth}
		\centering
		\caption{Michell cantilever beam}
		\includegraphics[width=\textwidth]{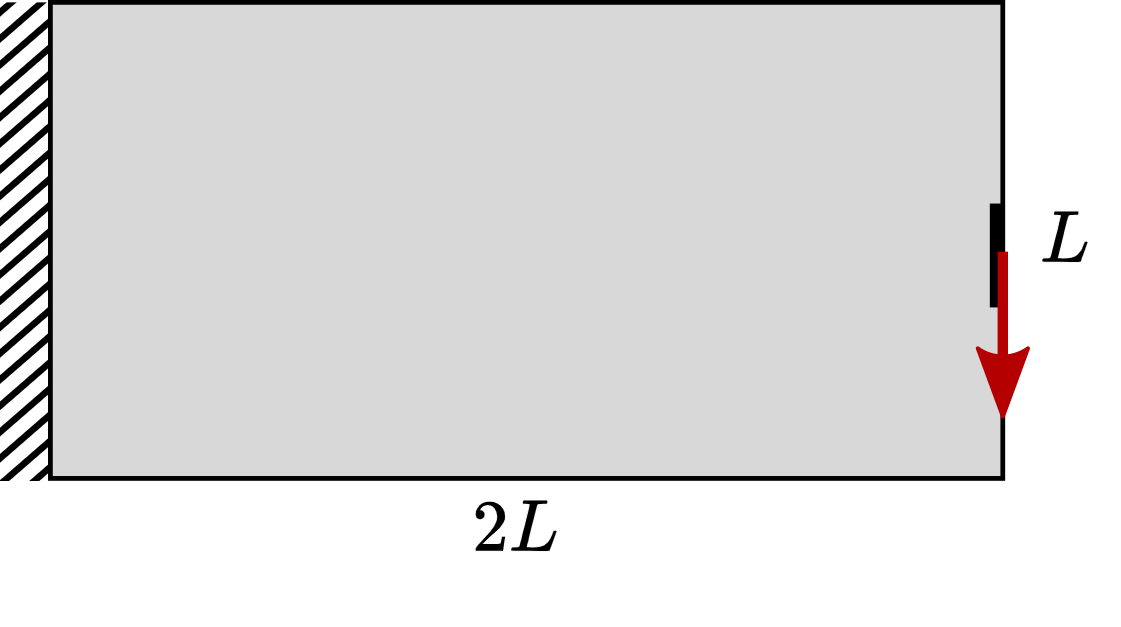}
		\label{fig:cantilever_BC}
	\end{subfigure}
	\hfill
	\begin{subfigure}[t]{0.34\textwidth}
		\centering
		\caption{L-shaped beam}
		\includegraphics[width=\textwidth]{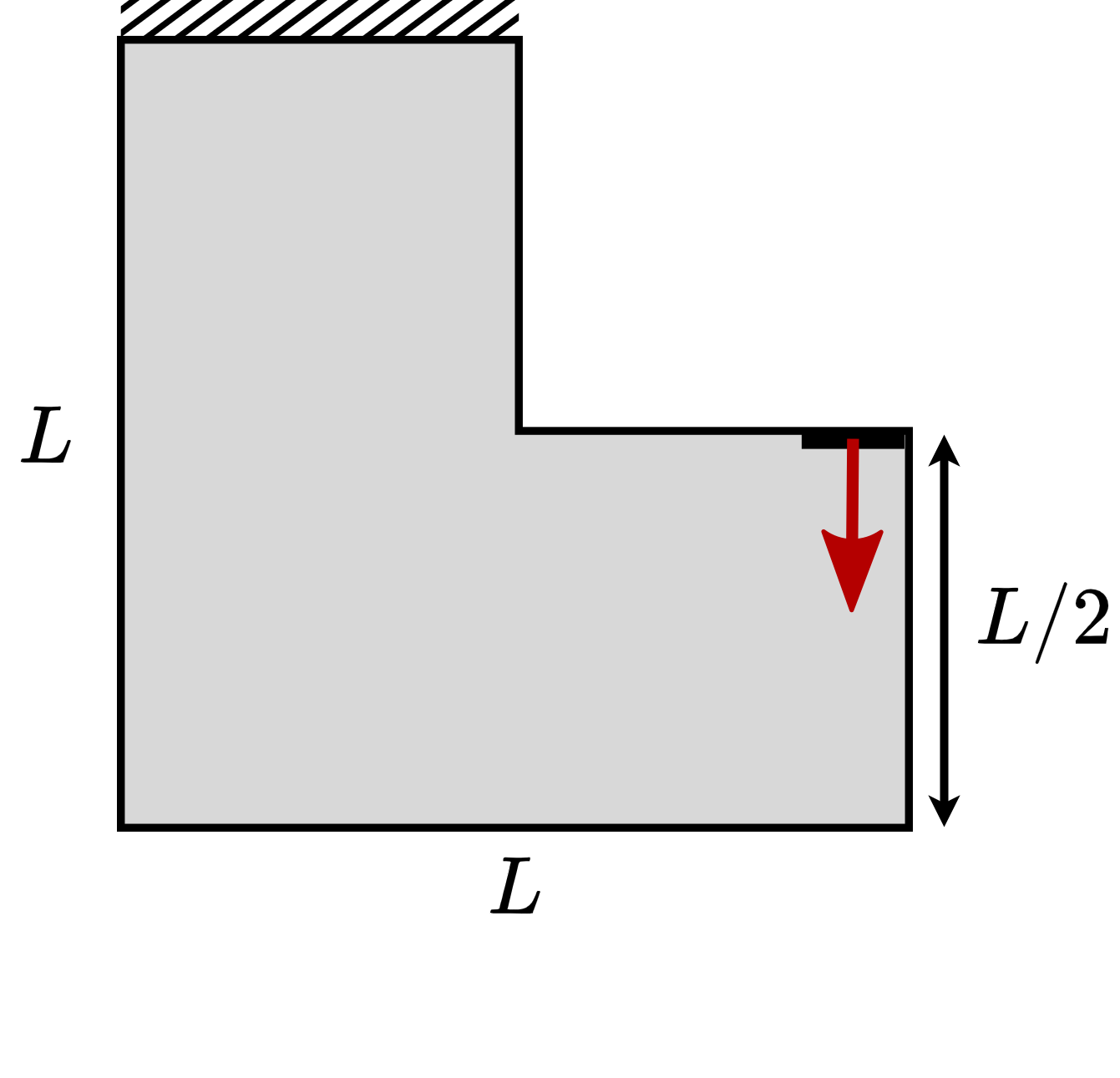}
		\label{fig:l_shaped_BC}
	\end{subfigure}
	\begin{subfigure}[t]{0.7\textwidth}
		\centering
		\caption{Double-clamped beam}
		\includegraphics[width=\textwidth]{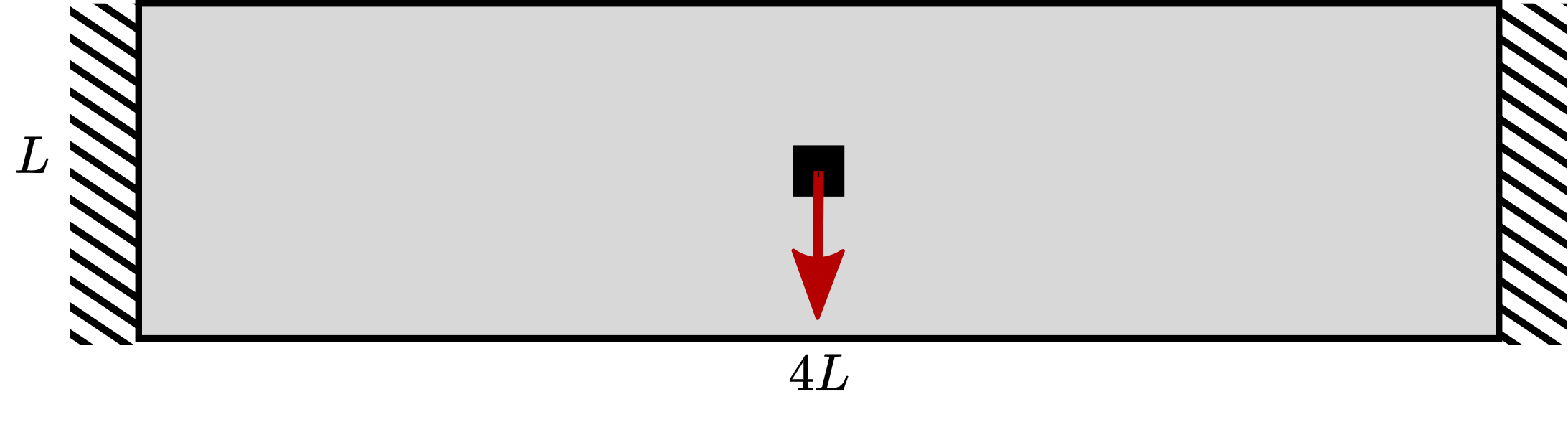}
		\label{fig:double_clamped_BC}
	\end{subfigure}
	\caption{Design domain and loading for the Michell cantilever, L-shaped and double-clamped beam}
	\label{fig:test_problems}
\end{figure}

The proposed method generalizes to a wide range of minimum compliance problems as the lamination fields and angles provided by the homogenization-based topology optimization method implicitly take boundary and load conditions into account. To demonstrate this capability the method is tested on three different problems; the Michell cantilever beam, the double-clamped beam and the L-shaped beam. See Figure \ref{fig:test_problems} for domain sizes and exact placement of boundaries and loads for the three cases. To avoid stress singularities at load locations the force is applied along a line of $\frac{1}{10}L$ in the Michell cantilever case, a line of $\frac{1}{20}L$ in the L-shaped beam case and on a block of $\frac{1}{10}L\times\frac{1}{10}L$ in the double-clamped beam case. Furthermore, all elements with a density of $\pmb{\rho}>0.99$ and in a small radius around the load are set to solid.
To generate the lamination widths $\pmb{\mu}_1,\pmb{\mu}_2$ used during the post-processing scheme and the angles $\pmb{\theta}$ serving as input for the neural network the optimization problem stated in eq. \ref{eq:opt_problem} is solved until convergence with $\gamma_{\theta}=0$, $\Gamma=0.05$, and a filter radius of $r_{min}=1.2$ for varying mesh sizes, volume fractions and relative minimum thicknesses $\mu_{min}$. For all of the numerical examples presented below replication padding with a width of 2 pixels has been added to the angular input for the neural network, to provide a better estimate of the image gradient near domain boundaries.

\subsection{Network training for specified periodicity}

Two different neural networks have been used to generate the results presented in the following sections; one trained for a periodicity equal to a wave-length of $\varepsilon_i=10 h_i$  and one for a wave-length of $\varepsilon_i=20 h_i$. Here $h_i$ refers to the element size on the intermediate mesh $\mathcal{T}_i$. Both networks have been trained using the two-step training procedure specified in Algorithm \ref{alg:two_step_train}. In the first step weight factors $\lambda_{\omega}=1$, $\lambda_{\tau}=1$ and $\lambda_b=0$ were used, while weight factors $\lambda_{\omega}=0$, $\lambda_{\tau}=1$ and $\lambda_{b}=2$ were used during the second step of training. In all cases the target total variation was set to $\tau=\left(2/\varepsilon_i\right)^2$, and the fixed size of the frequency band was set to $b=4$ pixels. For both the first and second step of training the Adam optimizer with an initial learning rate of $\eta_0=2\text{-e}4$, and $\beta_1=0.9$, $\beta_1=0.99$ was used. The network was trained for 10 epochs in the first step and 5 epochs in the second step.\\

A dataset consisting of 20.000 orientation fields sampled as described in Section \ref{sec:data+input} and encoded using 24 Gaussian channels was used for training.
The complete training process, i.e. application of Algorithm  \ref{alg:two_step_train}, for the full 15 epochs using this dataset took $\approx1$ hour on a NVIDIA Titan X GPU, and once trained a forward pass of the network takes only a few milli-seconds on a GPU. To demonstrate that the proposed method also can be used on a standard CPU all computation times presented in the following tables have been performed on a single Intel Core i7 2.6 GHz CPU. If a GPU is available significant speed-ups compared to the stated times can be expected, as the two most expensive operations of the post-processing scheme, the skeletonize and distance transform, can also be performed on a GPU. Wagner \cite{wagner2020a} reports a 50 times speed-up when performing skeletonization on the GPU, while Zampirolli et al. \cite{zampirolli2017a} report a factor five speed-up when performing Euclidean distance transform on the GPU.

\subsection{Michell cantilever beam}

The extensively studied Michell cantilever beam \citep{sigmund2016a} serves as good starting point for investigating the performance of the neural network-based de-homogenization approach for various parameter configurations. Table \ref{tab:michell_low_res_results} shows the results for a fixed homogenization-based resolution of $60\times30$ with varying periodicity, minimum relative thickness and volume fractions. The performance is measured in terms of compliance and compliance times volume fraction normalized by the corresponding reference value from the homogenization-based solution. Note that for small volume fractions, compliance and volume are close to inversely proportional. Consequently, we can use their product as an approximate means to compare structures with slightly varying volume fractions.

\rowcolors{2}{gray!25}{white}
\begin{table}[tb]
	\centering
	\begin{tabular}{c c c | c c | c c c c c | c}
	    \rowcolor{gray!50}
		$h_c$ & $\varepsilon_i$ & $\mu_{min}$ & $V_{ref}$ & $\mathcal{C}_{ref}$ & $h_f$ & $\varepsilon_f$ & $V_f$ & $\mathcal{C}_f$ & $\frac{\mathcal{C}_{f} \cdot V_{f}}{\mathcal{C}_{ref} \cdot V_{ref}}$ & $t_f$[s]  \\ \hline
		1/30 & $20h_i$ & 0.05 & 0.2535 & 106.21 & $1/24 h_c$ & $60 h_f$ & 0.2695 & 140.16 & 1.4030 & 1.28 \\
		1/30 & $20h_i$ & 0.05 & 0.4024 & 68.58  & $1/24 h_c$ & $60 h_f$  & 0.4329 & 75.25  & 1.1804 & 1.27  \\
		1/30 & $20h_i$ & 0.10 & 0.2568 & 113.61  & $1/24 h_c$ & $60 h_f$  & 0.2661 & 167.32 & 1.5260 & 1.26  \\
		1/30 & $20h_i$ & 0.10 & 0.4080 & 69.00  & $1/24 h_c$ & $60 h_f$  & 0.4393 & 73.68 & 1.1499 & 1.33 \\
		1/30 & $20h_i$ & 0.20 & 0.2614 & 122.86  & $1/24 h_c$ & $60 h_f$  & 0.2665 & 141.57 & 1.1747 & 1.28  \\
		1/30 & $20h_i$ & 0.20 & 0.4165 & 73.37  &  $1/24 h_c$ & $60 h_f$  & 0.4389 & 77.07 &  1.1070 & 1.28  \\
		1/30 & $10h_i$ & 0.05 & 0.2535 & 106.21  & $1/40 h_c$ & $50 h_f$  & 0.2581 & 143.65 & 1.3770 & 2.23  \\
		1/30 & $10h_i$ & 0.05 & 0.4024 & 68.58  & $1/40 h_c$ & $50 h_f$  & 0.4201 & 78.79 &1.1993 & 2.18  \\
		1/30 & $10h_i$ & 0.10 & 0.2568 & 113.61  & $1/40 h_c$ & $50 h_f$  & 0.2569 & 149.53 & 1.3167 & 2.10  \\
		1/30 & $10h_i$ & 0.10 & 0.4080 & 69.00  & $1/40 h_c$ & $50 h_f$  & 0.4224 & 75.16 & 1.1279 & 2.19  \\
		1/30 & $10h_i$ & 0.20 & 0.2614 & 122.86  & $1/40 h_c$ & $50 h_f$  & 0.2566 & 152.86 & 1.2214 & 2.21  \\
		1/30 & $10h_i$ & 0.20 & 0.4165 & 73.37  &  $1/40 h_c$ & $50 h_f$  & 0.4323 & 77.39 & 1.0949 & 2.16  \\
	\end{tabular}
	\caption{Performance and computational cost of neural network based de-homogenization approach on the Michell cantilever beam for a $60\times30$ input. Here $h_c$ is the element size on the coarse mesh (homogenization mesh), $\varepsilon_i$ is the wave-length on the intermediate mesh (output of neural network), $\mu_{min}$ is the minimum relative thickness, $V_{ref}$ is the reference volume fraction from the homogenization solution, $\mathcal{C}_{ref}$ is the reference compliance from the homogenization solution, $h_f$ is the element size on the fine mesh (de-homogenization mesh), $\varepsilon_f$ is the wave-length on the fine mesh, $V_f$ is the volume fraction of the de-homogenized design, $\mathcal{C}_f$ is the compliance of the de-homogenized design, $\frac{\mathcal{C}_{f} \cdot V_{f}}{\mathcal{C}_{ref} \cdot V_{ref}}$ is the ratio between the performance of the de-homogenized design and the reference solution (lower is better), and $t_f$ is the computation time for the de-homogenization in seconds. }
	\label{tab:michell_low_res_results}
\end{table}
\rowcolors{2}{white}{white}

\begin{figure}[tb]
	\centering
	\begin{subfigure}[t]{0.32\textwidth}
		\centering
		\includegraphics[width=\textwidth]{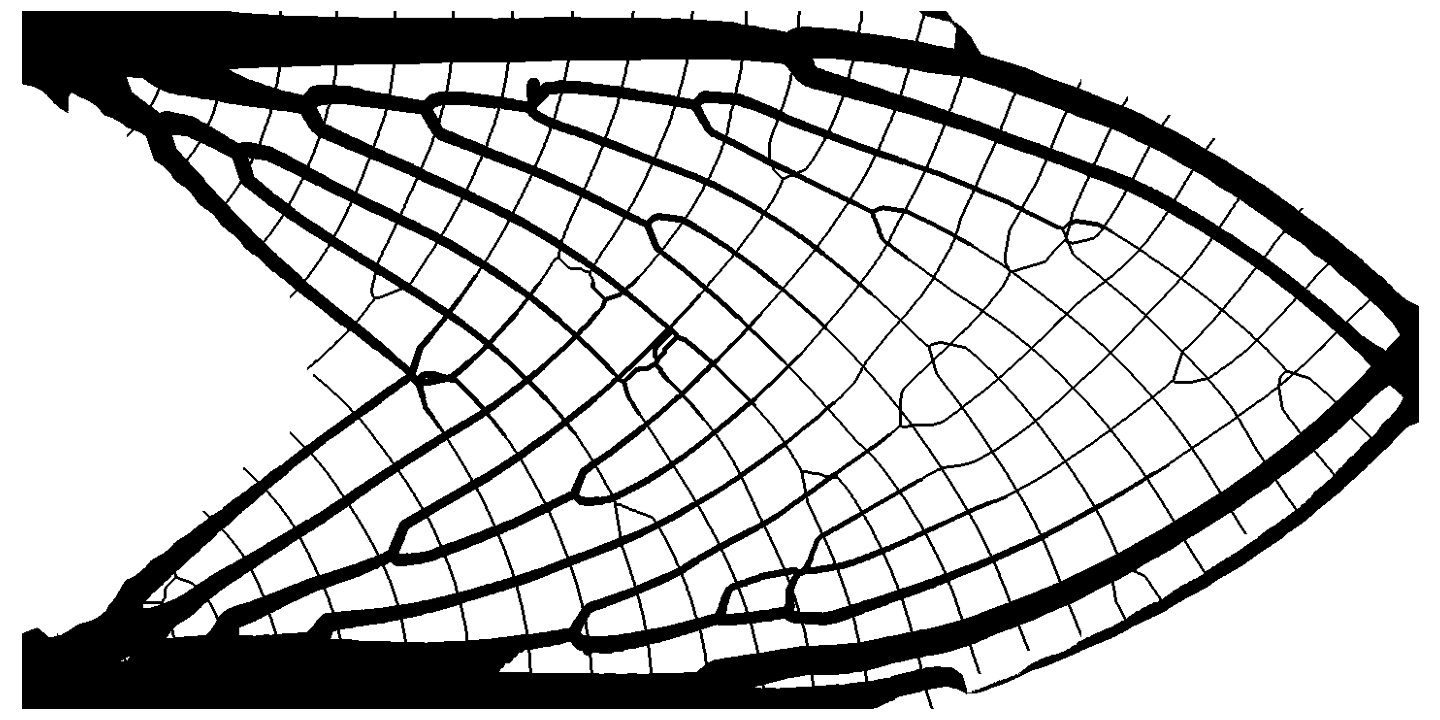}
		\caption{$V=0.25$, $\mu_{min}=0.05$}
		\label{fig:michell_60_30_vol_0.25_MinMu_0.05_p20}
	\end{subfigure}
	\begin{subfigure}[t]{0.32\textwidth}
		\centering
		\includegraphics[width=\textwidth]{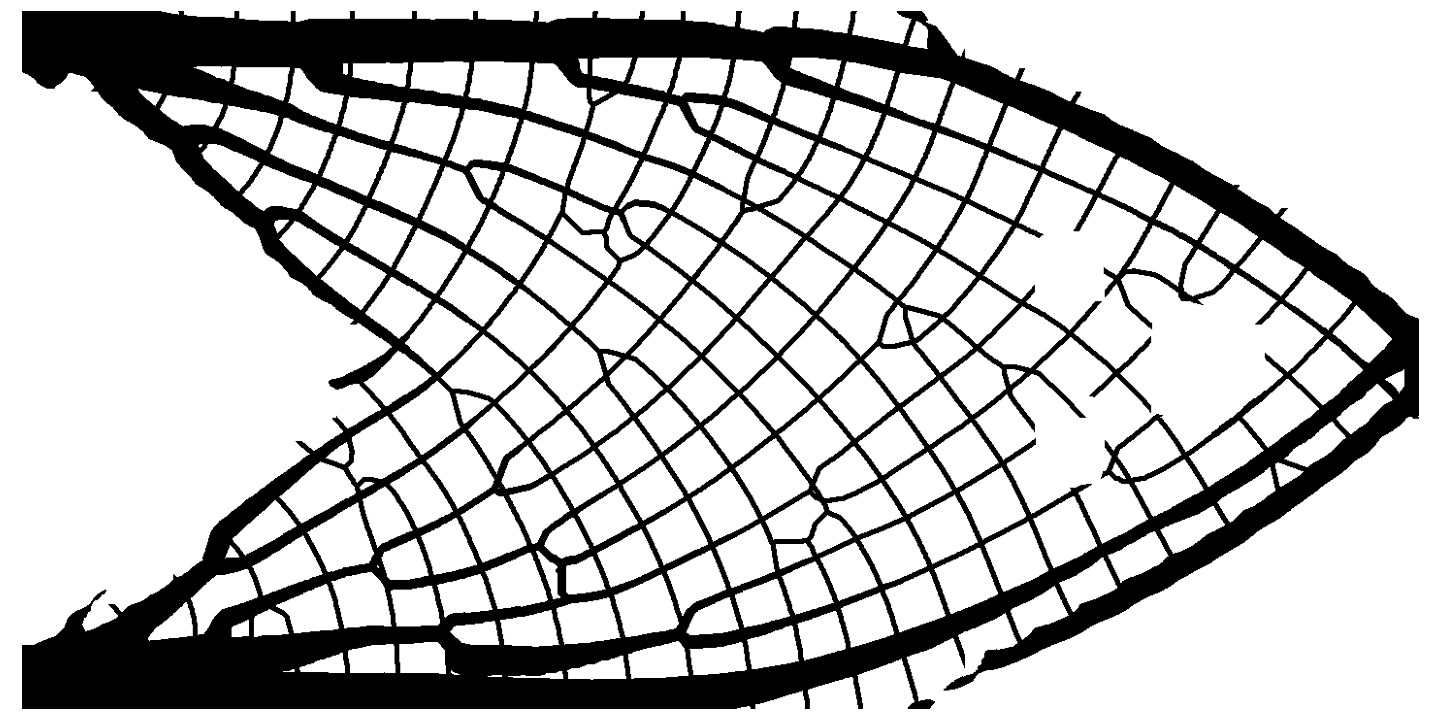}
		\caption{$V=0.25$, $\mu_{min}=0.10$}
		\label{fig:michell_60_30_vol_0.25_MinMu_0.10_p20}
	\end{subfigure}
	\begin{subfigure}[t]{0.32\textwidth}
		\centering
		\includegraphics[width=\textwidth]{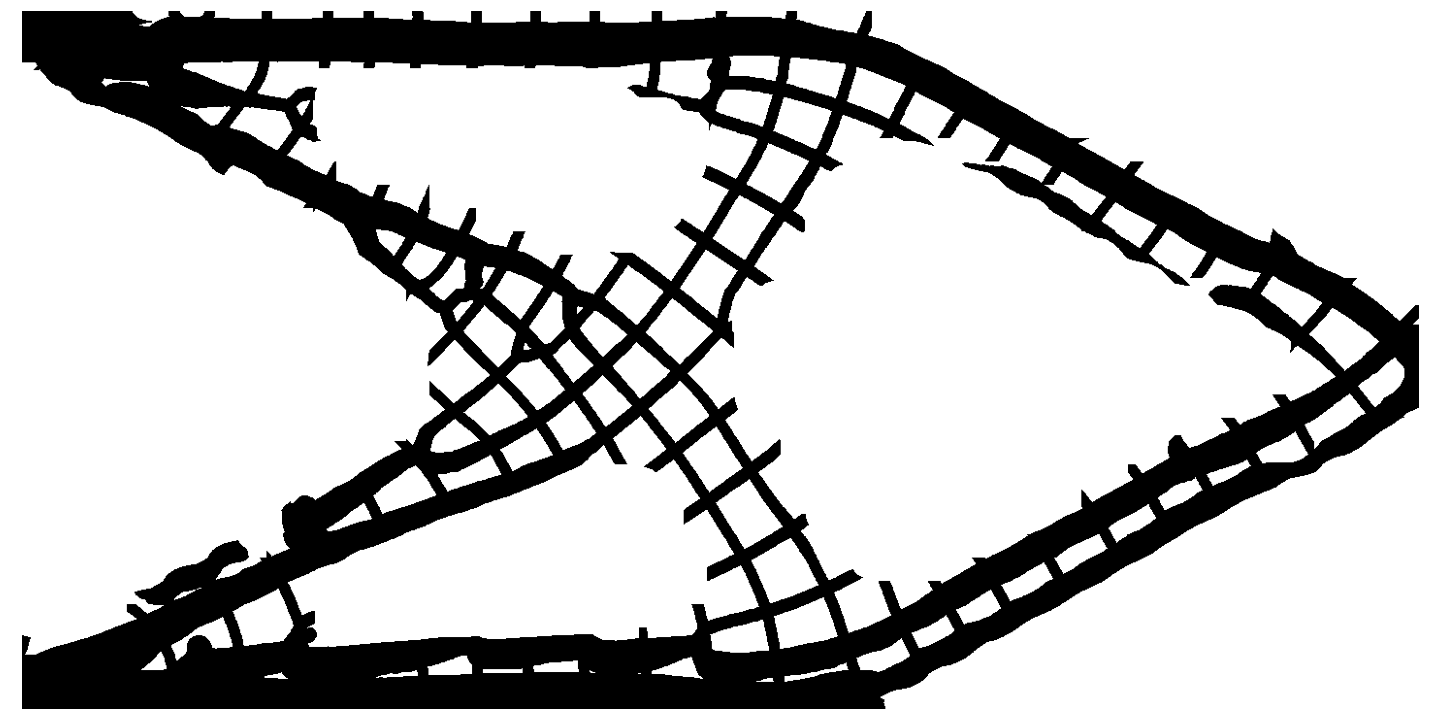}
		\caption{$V=0.25$, $\mu_{min}=0.20$}
		\label{fig:michell_60_30_vol_0.25_MinMu_0.20_p20}
	\end{subfigure}
	\begin{subfigure}[t]{0.32\textwidth}
		\centering
		\includegraphics[width=\textwidth]{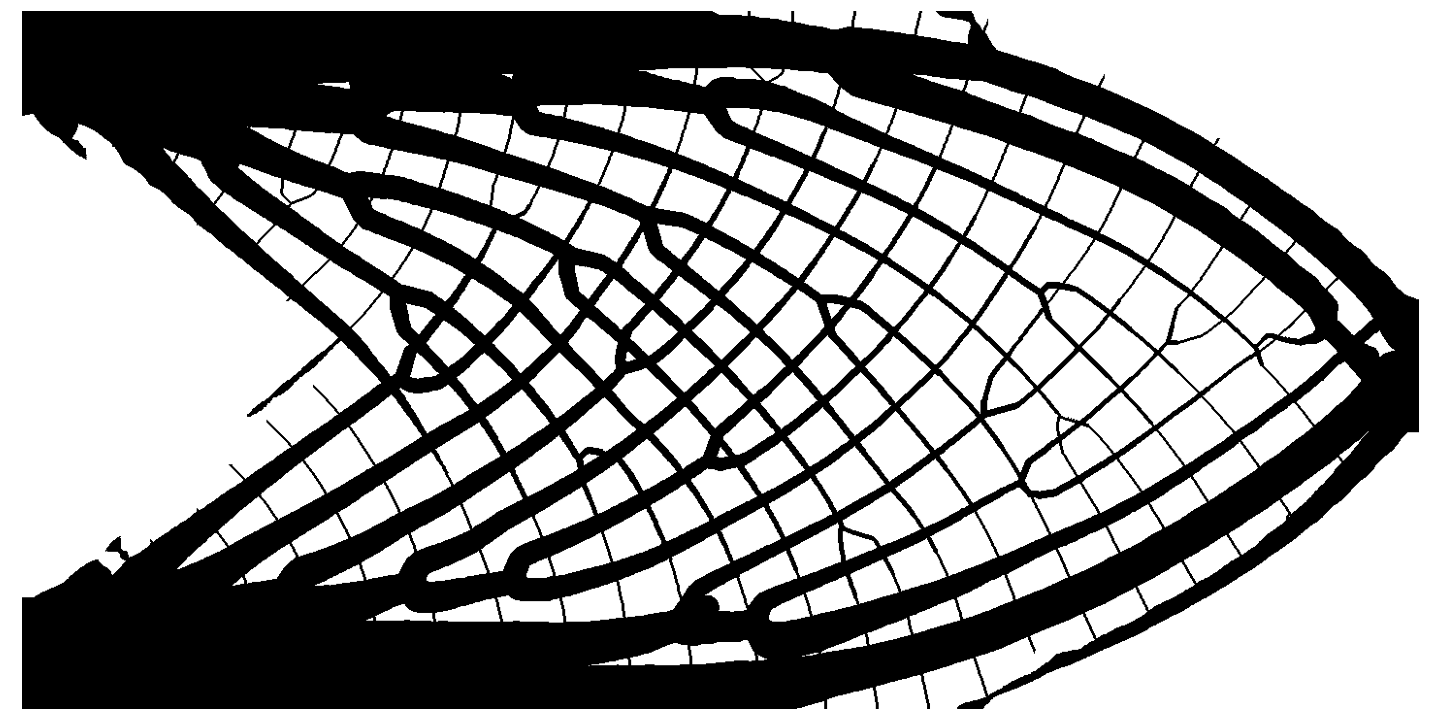}
		\caption{$V=0.40$, $\mu_{min}=0.05$}
		\label{fig:michell_60_30_vol_0.40_MinMu_0.05_p20}
	\end{subfigure}
	\begin{subfigure}[t]{0.32\textwidth}
		\centering
		\includegraphics[width=\textwidth]{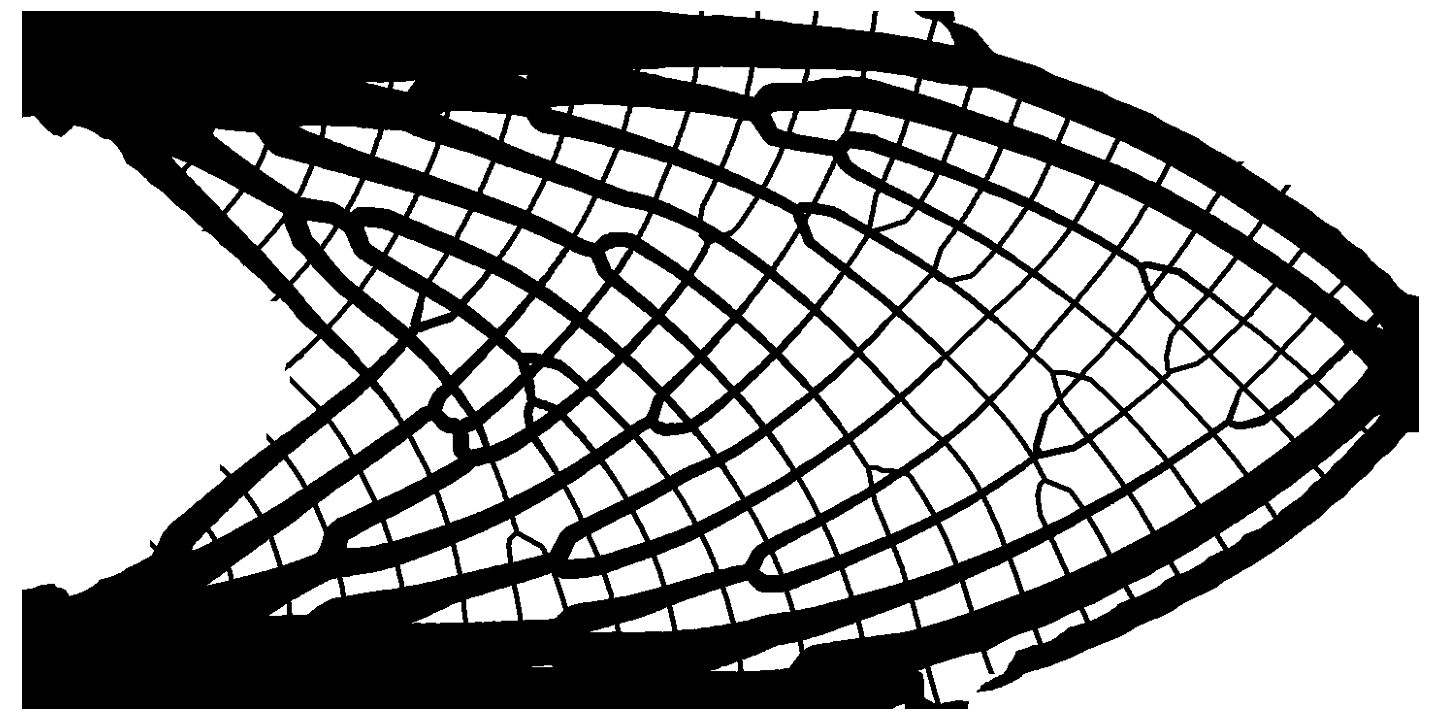}
		\caption{$V=0.40$, $\mu_{min}=0.10$}
		\label{fig:michell_60_30_vol_0.40_MinMu_0.10_p20}
	\end{subfigure}
	\begin{subfigure}[t]{0.32\textwidth}
		\centering
		\includegraphics[width=\textwidth]{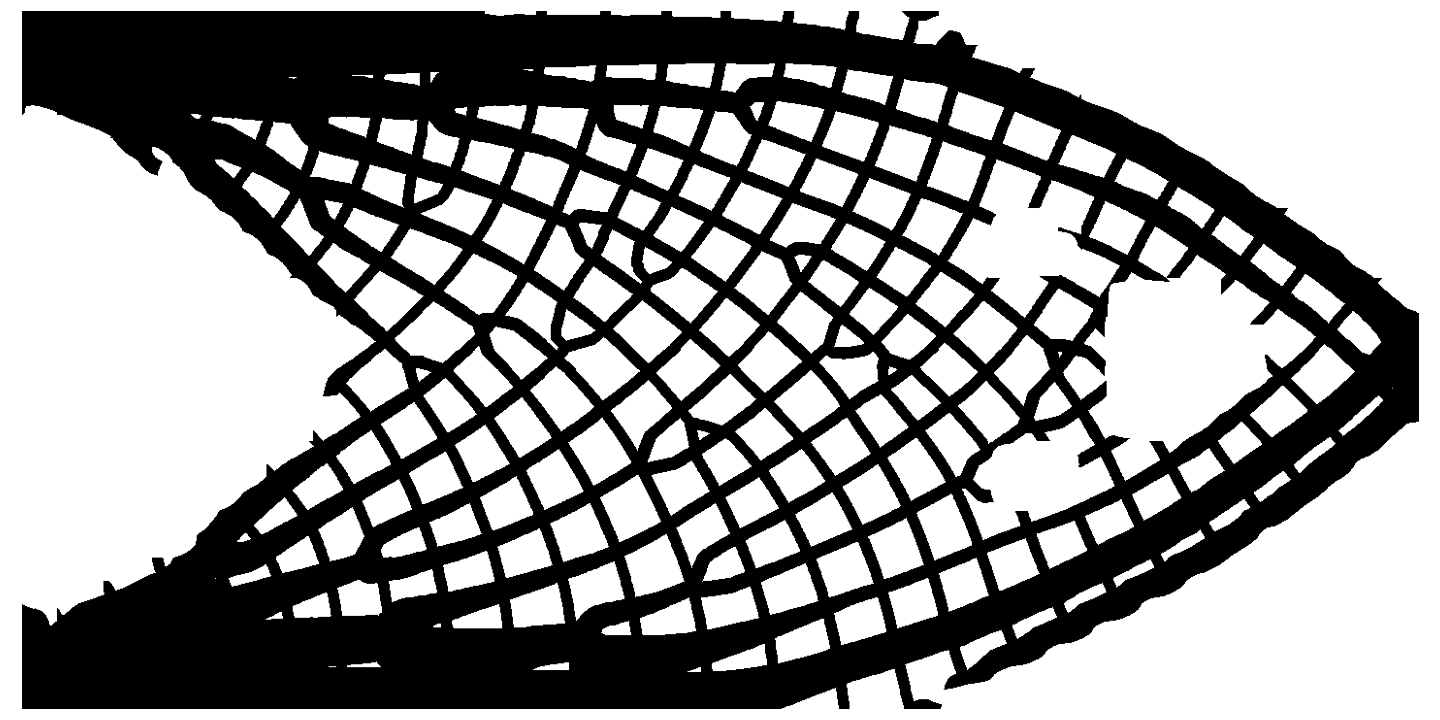}
		\caption{$V=0.40$, $\mu_{min}=0.20$}
		\label{fig:michell_60_30_vol_0.40_MinMu_0.20_p20}
	\end{subfigure}
	\caption{De-homogenization of a $60\times30$ Michell cantilever input to a fine mesh of $1440\times720$ elements with a wave-length of $\varepsilon_f=60h_f$.}
	\label{fig:michell_low_res_p20}
\end{figure}

\begin{figure}[tb]
	\centering
	\begin{subfigure}[t]{0.32\textwidth}
		\centering
		\includegraphics[width=\textwidth]{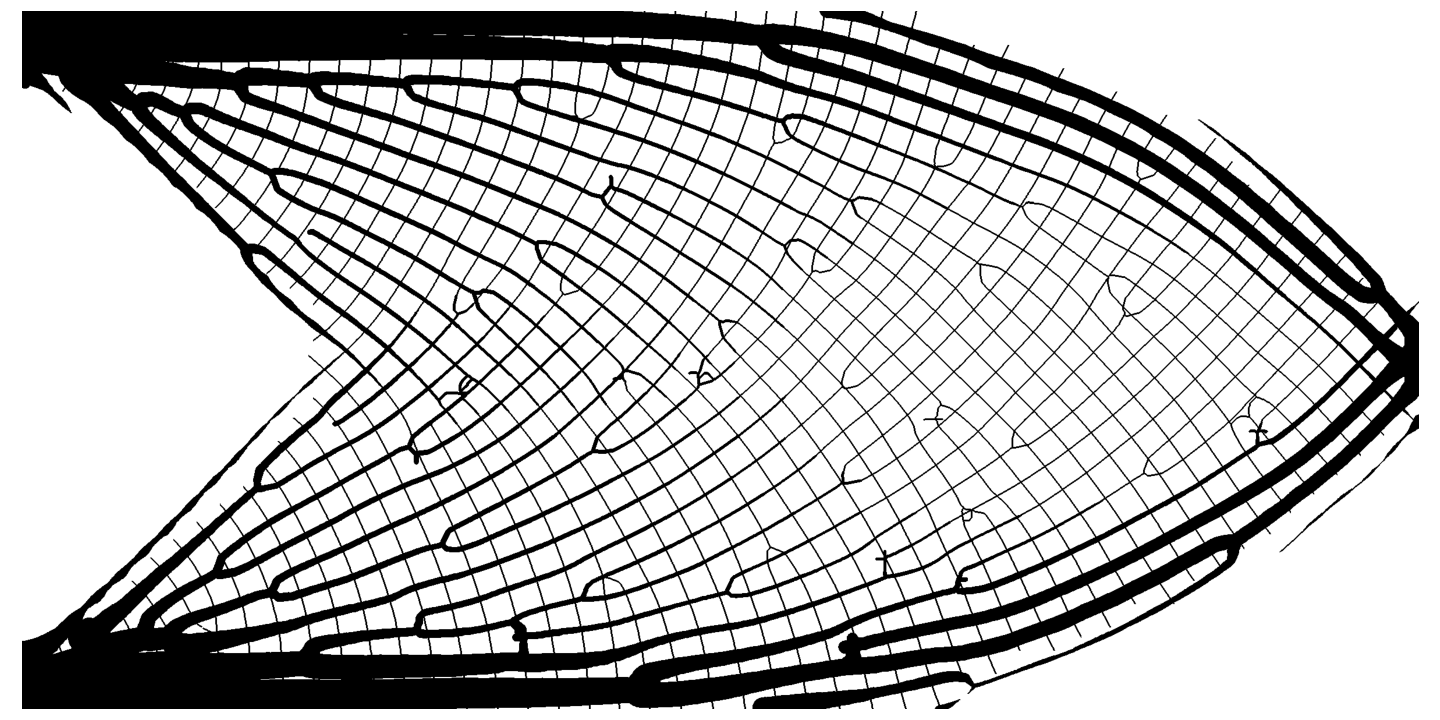}
		\caption{$V=0.25$, $\mu_{min}=0.05$}
		\label{fig:michell_60_30_vol_0.25_MinMu_0.05_p10}
	\end{subfigure}
	\begin{subfigure}[t]{0.32\textwidth}
		\centering
		\includegraphics[width=\textwidth]{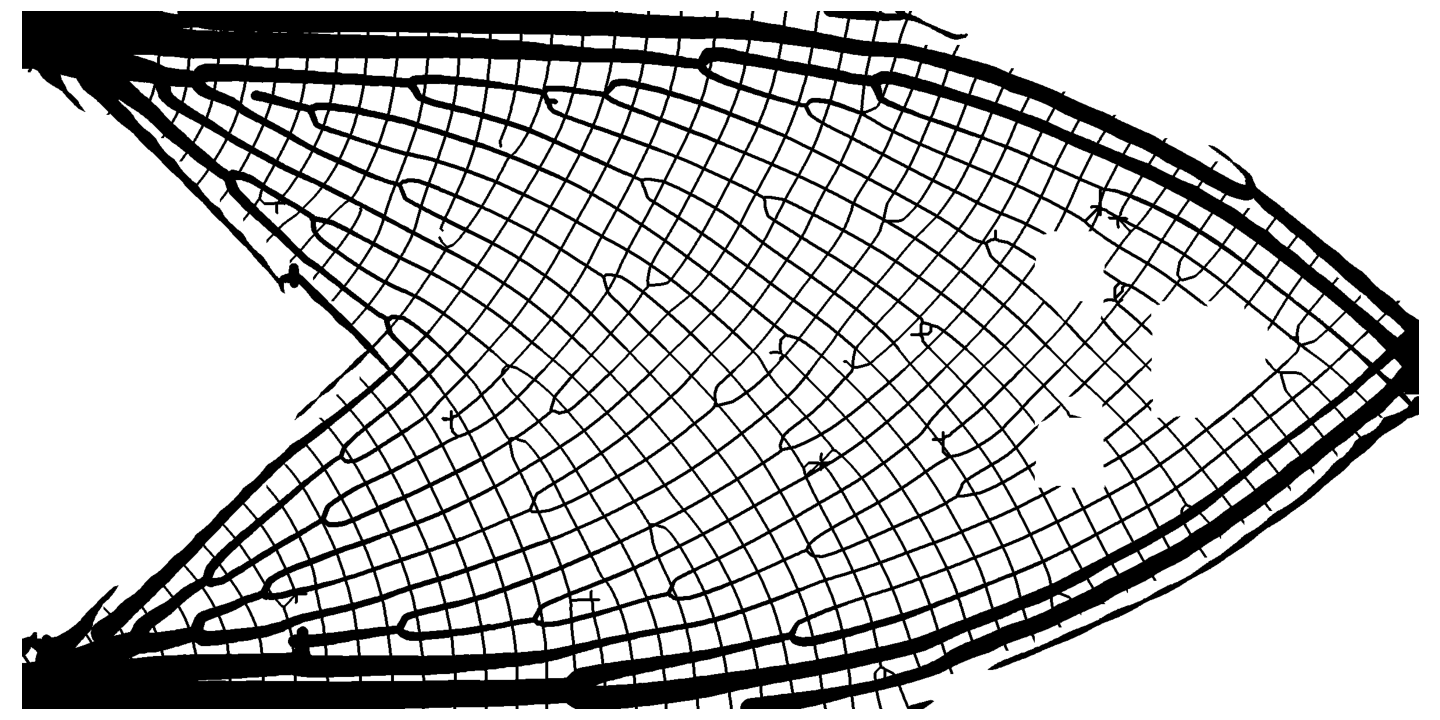}
		\caption{$V=0.25$, $\mu_{min}=0.10$}
		\label{fig:michell_60_30_vol_0.25_MinMu_0.10_p10}
	\end{subfigure}
	\begin{subfigure}[t]{0.32\textwidth}
		\centering
		\includegraphics[width=\textwidth]{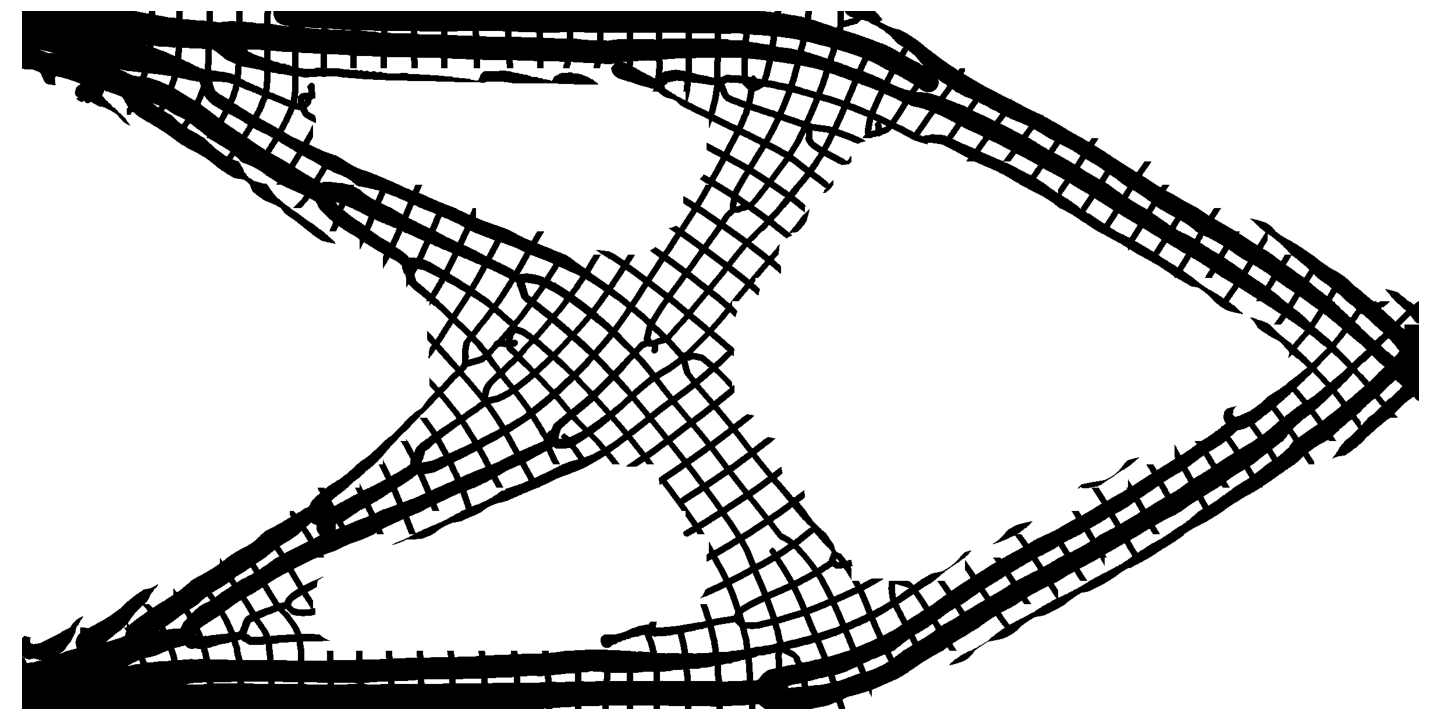}
		\caption{$V=0.25$, $\mu_{min}=0.20$}
		\label{fig:michell_60_30_vol_0.25_MinMu_0.20_p10}
\end{subfigure}
	\begin{subfigure}[t]{0.32\textwidth}
		\centering
		\includegraphics[width=\textwidth]{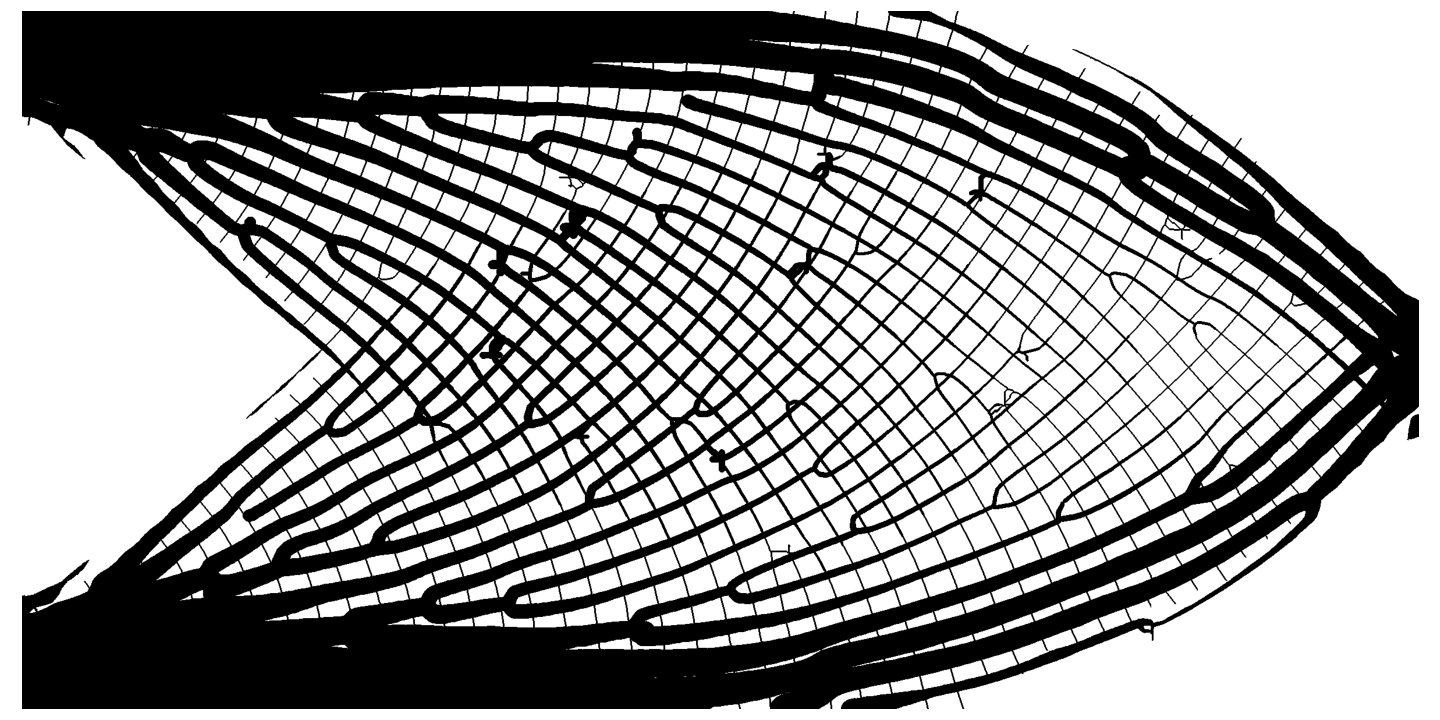}
		\caption{$V=0.40$, $\mu_{min}=0.05$}
		\label{fig:michell_60_30_vol_0.40_MinMu_0.05_p10}
	\end{subfigure}
	\begin{subfigure}[t]{0.32\textwidth}
		\centering
		\includegraphics[width=\textwidth]{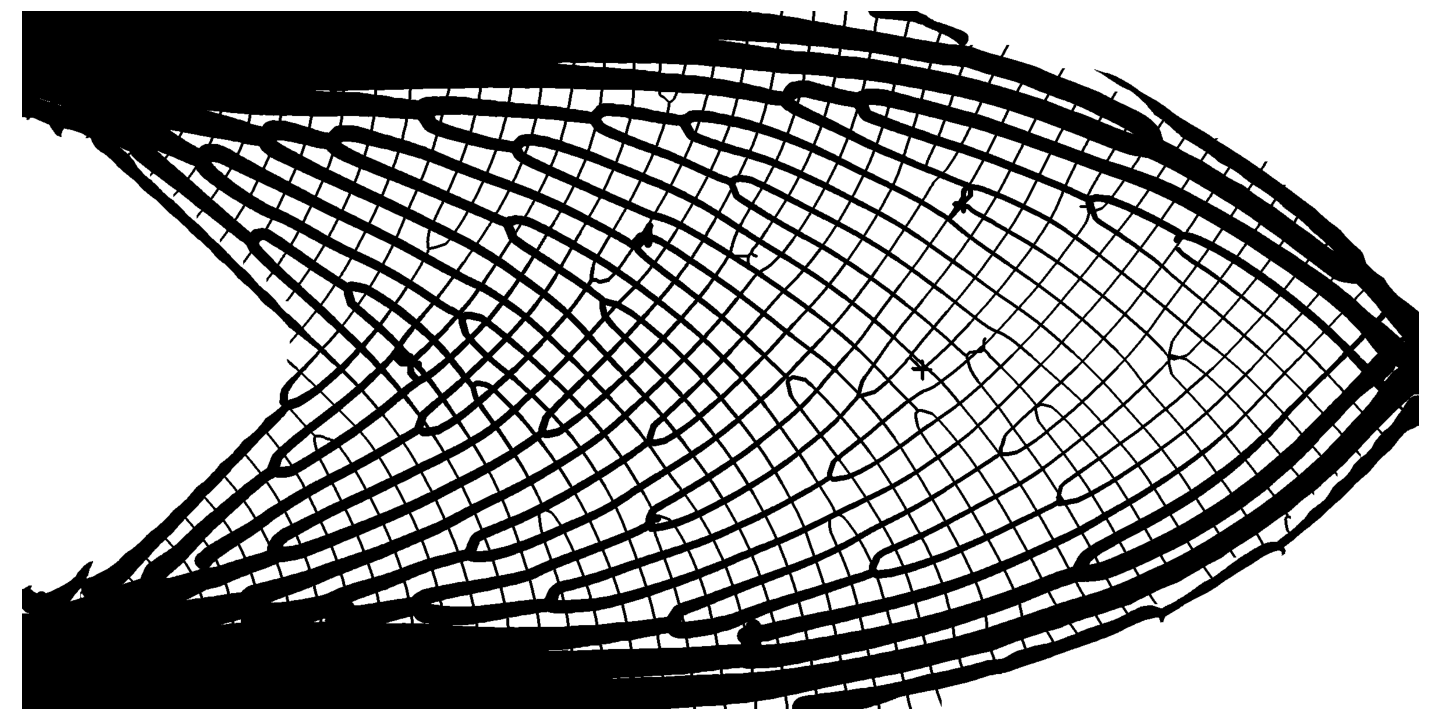}
		\caption{$V=0.40$, $\mu_{min}=0.10$}
		\label{fig:michell_60_30_vol_0.40_MinMu_0.10_p10}
	\end{subfigure}
	\begin{subfigure}[t]{0.32\textwidth}
		\centering
		\includegraphics[width=\textwidth]{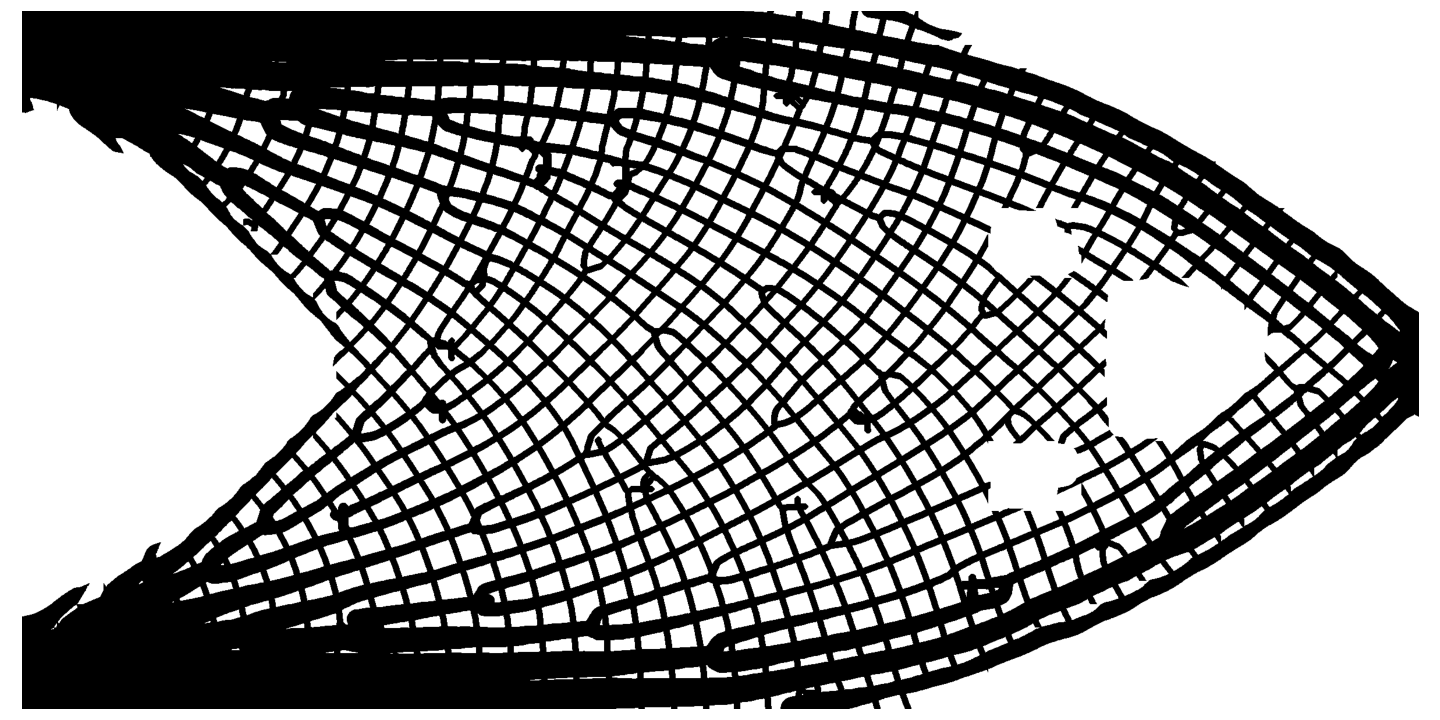}
		\caption{$V=0.40$, $\mu_{min}=0.20$}
		\label{fig:michell_60_30_vol_0.40_MinMu_0.20_p100}
	\end{subfigure}
	\caption{De-homogenization of a $60\times30$ Michell cantilever input to a fine mesh of $2400\times1200$ elements with a wave-length of $\varepsilon_f=50h_f$.}
	\label{fig:michell_low_res_p10}
\end{figure}

Despite the proposed method not relying on any physical (finite element) modeling in either training or post-processing, the de-homogenized designs are found to perform very well when compared to the reference homogenization-based result. Moreover, the discrepancy is consistently reduced as the wave-length is decreased. That is, on average the de-homogenized solutions performs around $25.7\%$ worse in terms of $C _f\cdot V_f$ for $\varepsilon_i=20h_i$, and $22.3\%$ worse for $\varepsilon_i=10h_i$. It is also observed that the proposed method performs best for high volume fractions, which in part can be explained by the fact that structural member thicknesses are better resolved as the volume increases but also because the stiffness of the resulting sub-optimal curved forks is much higher for higher local volume fractions.\\
Quantitatively, the periodicity does not seem to affect the performance significantly, and thus one could argue that using a larger wave-length should be preferred, as the minimum relative thickness can be resolved on a coarser mesh in this case. However, visual inspection of the designs indicates that the lower wave-length designs seem more robust, for example the diagonal bars in Figure \ref{fig:michell_60_30_vol_0.25_MinMu_0.20_p20} are only resolved by one period in the high wave-length case, whereas multiple periods are present in the low wave-length case, c.f. Figure \ref{fig:michell_60_30_vol_0.25_MinMu_0.20_p10}. Boundary artefacts in the form of non-load carrying material is present in many of the designs, especially for larger wave-lengths. One idea to get rid of these could be to use the scheme proposed by Groen et al. \cite{groen2018a}, in which finite element analyses are used to iteratively identify non-load carrying material, and set these elements to void. Incorporating this post-processing procedure could help reduce the volume fraction used by the neural network-based de-homogenization approach, as it generally tends to overshoot the reference volume fraction slightly, i.e. $5.7\%$ on average for the large wave-length case, and $2.4\%$ for the small wave-length case.  \\

Next, a higher resolution homogenization-based topology optimization result is used as input to the neural network. The results for a $240\times120$ input de-homogenized onto a $5760\times2880$ mesh are shown in Table \ref{tab:michell_high_res_results}. Here one would expect a higher performance than in the low-resolution input case as the homogenization-based solution itself is of higher quality and individual sub-optimal branches should have less impact on final de-homogenized design. Compared to the low-resolution input case a significant improvement in performance can be observed with the average performance in terms of $C _f\cdot V_f$  being within $13\%$ of the reference value. However, this comes at increased CPU cost as discussed next.

\rowcolors{2}{gray!25}{white}
\begin{table}[tb]
	\centering
	\begin{tabular}{c c c | c c | c c c c c | c}
		\rowcolor{gray!50}
		$h_c$ & $\varepsilon_i$ &  $\mu_{min}$ & $V_{ref}$ & $\mathcal{C}_{ref}$ & $h_f$ & $\varepsilon_f$ & $V_f$ & $\mathcal{C}_f$ & $\frac{\mathcal{C}_{f} \cdot V_{f}}{\mathcal{C}_{ref} \cdot V_{ref}}$  & $t_f$[s] \\ \hline
		1/120 & $20h_i$ & 0.05 & 0.2532 & 105.98 & $1/24 h_c$ & $60 h_f$ & 0.2587 & 130.40 & 1.2575 & 23.30 \\
		1/120 & $20h_i$  & 0.05 & 0.4023 & 68.61 & $1/24 h_c$ & $60 h_f$ & 0.4216& 73.97 & 1.1297 & 23.47 \\
		1/120 & $20h_i$  & 0.10 & 0.2566 & 111.55 & $1/24 h_c$ & $60 h_f$ & 0.2586 & 128.85 & 1.1642 & 23.41 \\
		1/120 & $20h_i$  & 0.10 & 0.4078 & 68.94 & $1/24 h_c$ & $60 h_f$ & 0.4237 & 72.25 & 1.0885 & 23.71 \\
		1/120 & $20h_i$  & 0.20 & 0.2572 & 113.19 & $1/24 h_c$ & $60 h_f$ & 0.2607 & 120.00 & 1.0745 & 24.02 \\
		1/120 & $20h_i$  & 0.20 & 0.4154 & 71.86 & $1/24 h_c$ & $60 h_f$ & 0.4323 & 73.72 & 1.0677 &  23.55 \\
	\end{tabular}
	\caption{Performance and computational cost of neural network based de-homogenization approach on the Michell cantilever beam for a high-resolution input.}
	\label{tab:michell_high_res_results}
\end{table}
\rowcolors{2}{white}{white}

\begin{figure}[tb]
	\centering
	\begin{subfigure}[t]{0.48\textwidth}
		\centering
		\includegraphics[width=\textwidth]{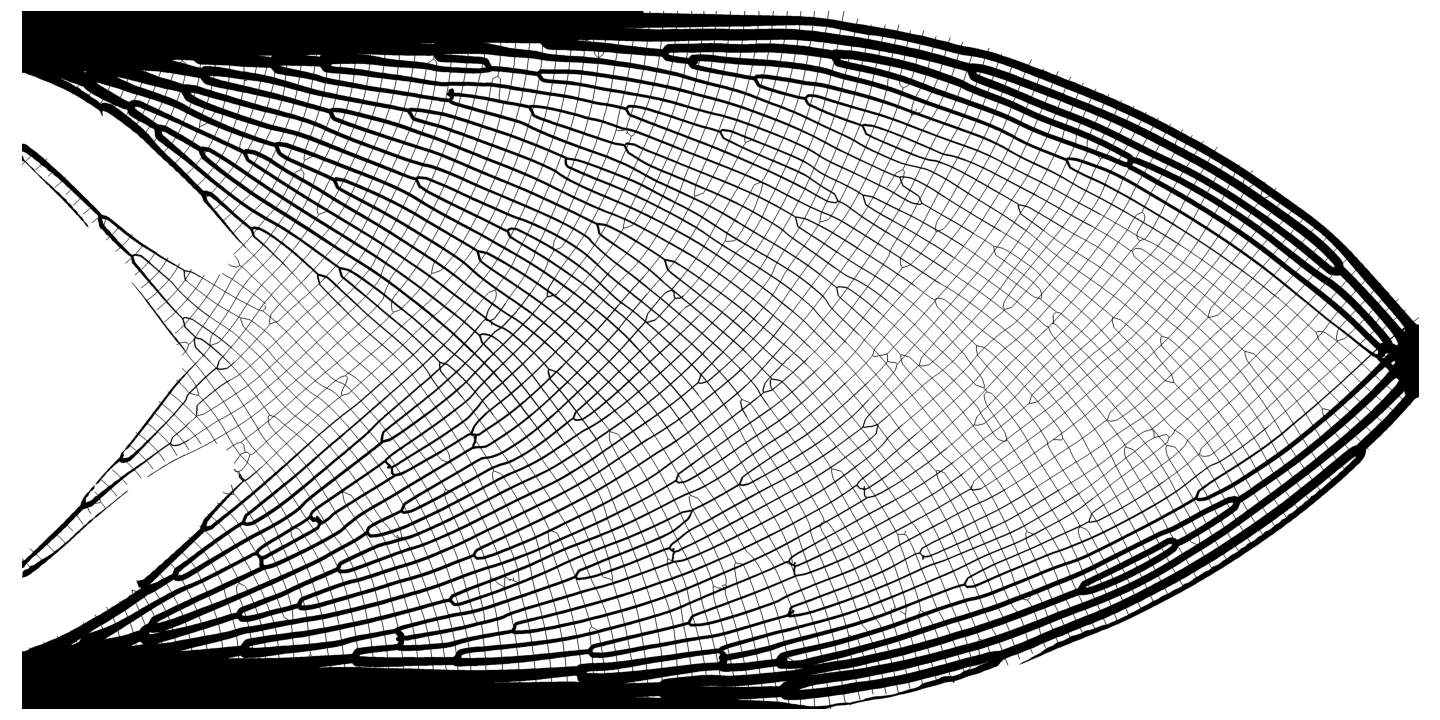}
		\caption{$V=0.25$, $\mu_{min}=0.05$}
		\label{fig:michell_240_120_vol_0.25_MinMu_0.05_p20}
	\end{subfigure}
	\hfill
	\begin{subfigure}[t]{0.48\textwidth}
		\centering
		\includegraphics[width=\textwidth]{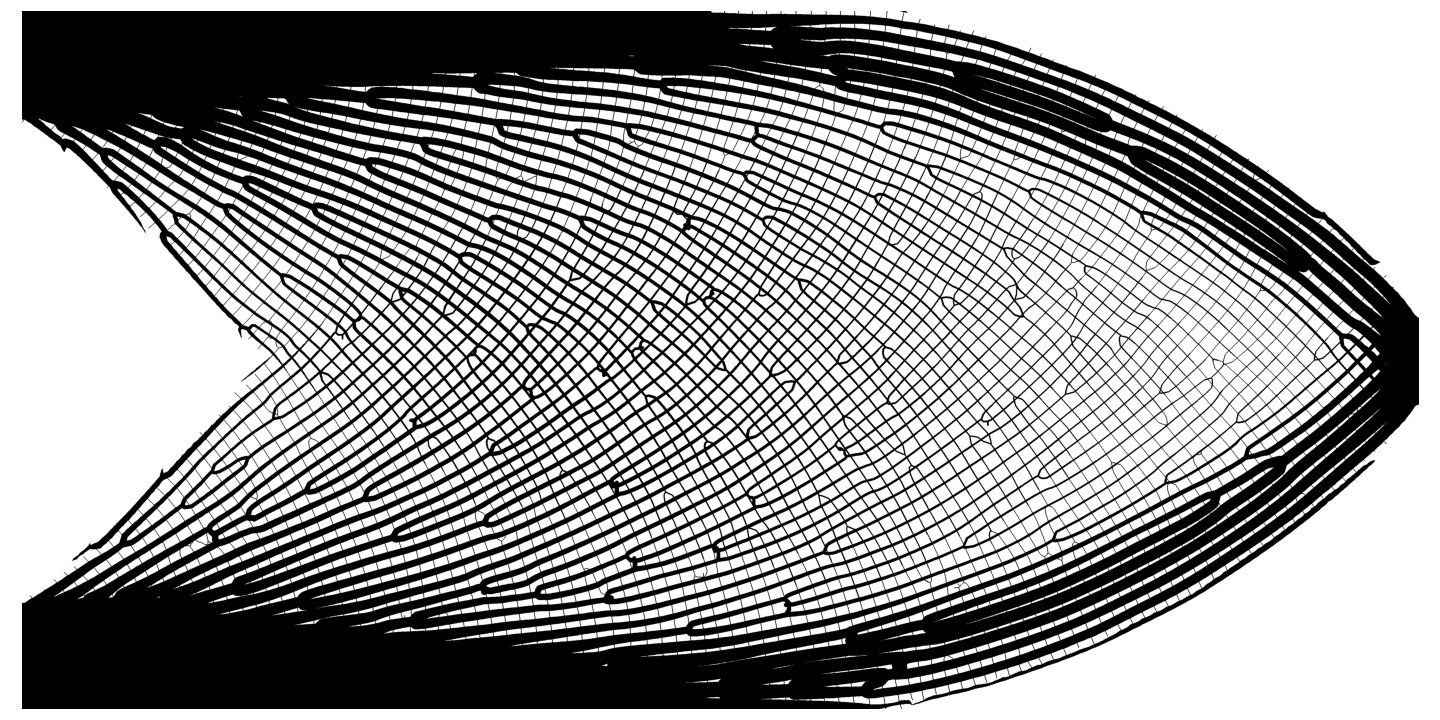}
		\caption{$V=0.40$, $\mu_{min}=0.05$}
		\label{fig:michell_240_120_vol_0.40_MinMu_0.05_p20}
	\end{subfigure}
	\begin{subfigure}[t]{0.48\textwidth}
		\centering
		\includegraphics[width=\textwidth]{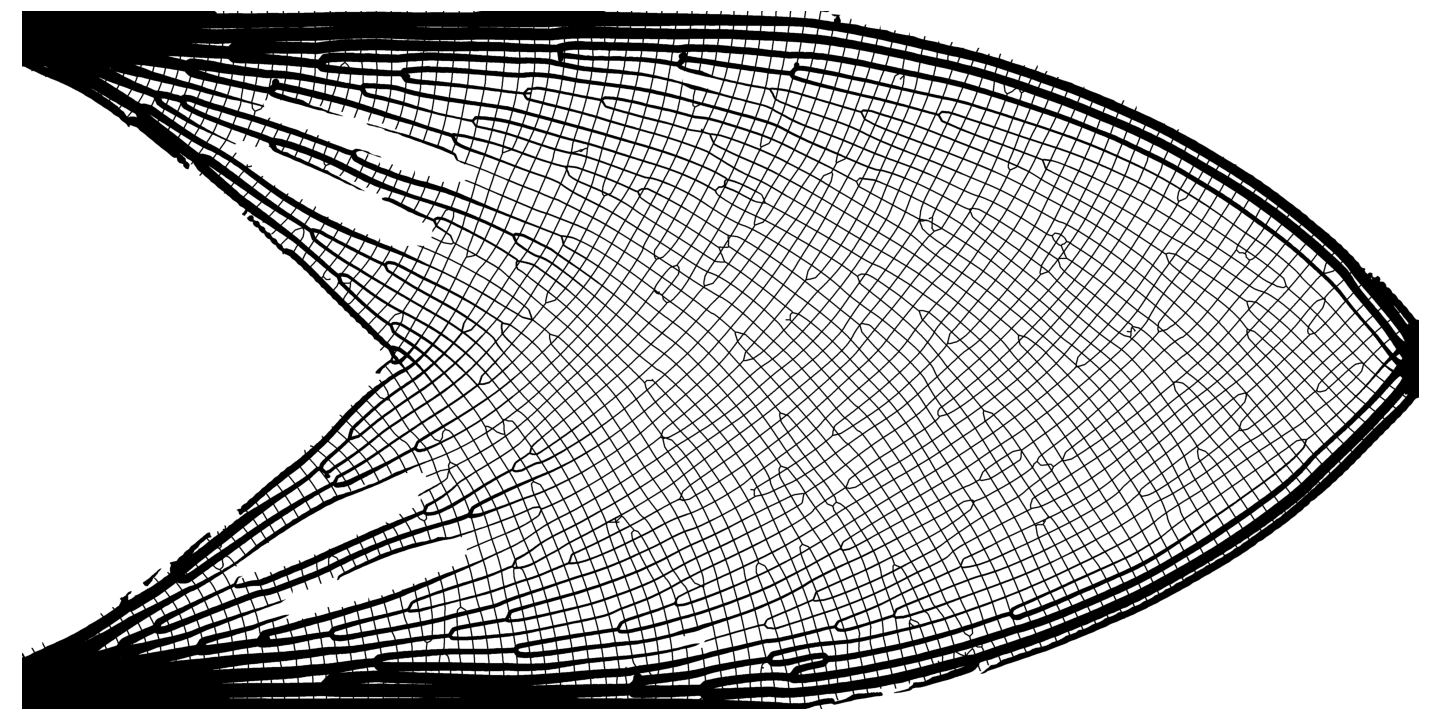}
		\caption{$V=0.25$, $\mu_{min}=0.10$}
		\label{fig:michell_240_120_vol_0.25_MinMu_0.10_p20}
	\end{subfigure}
	\hfill
	\begin{subfigure}[t]{0.48\textwidth}
		\centering
		\includegraphics[width=\textwidth]{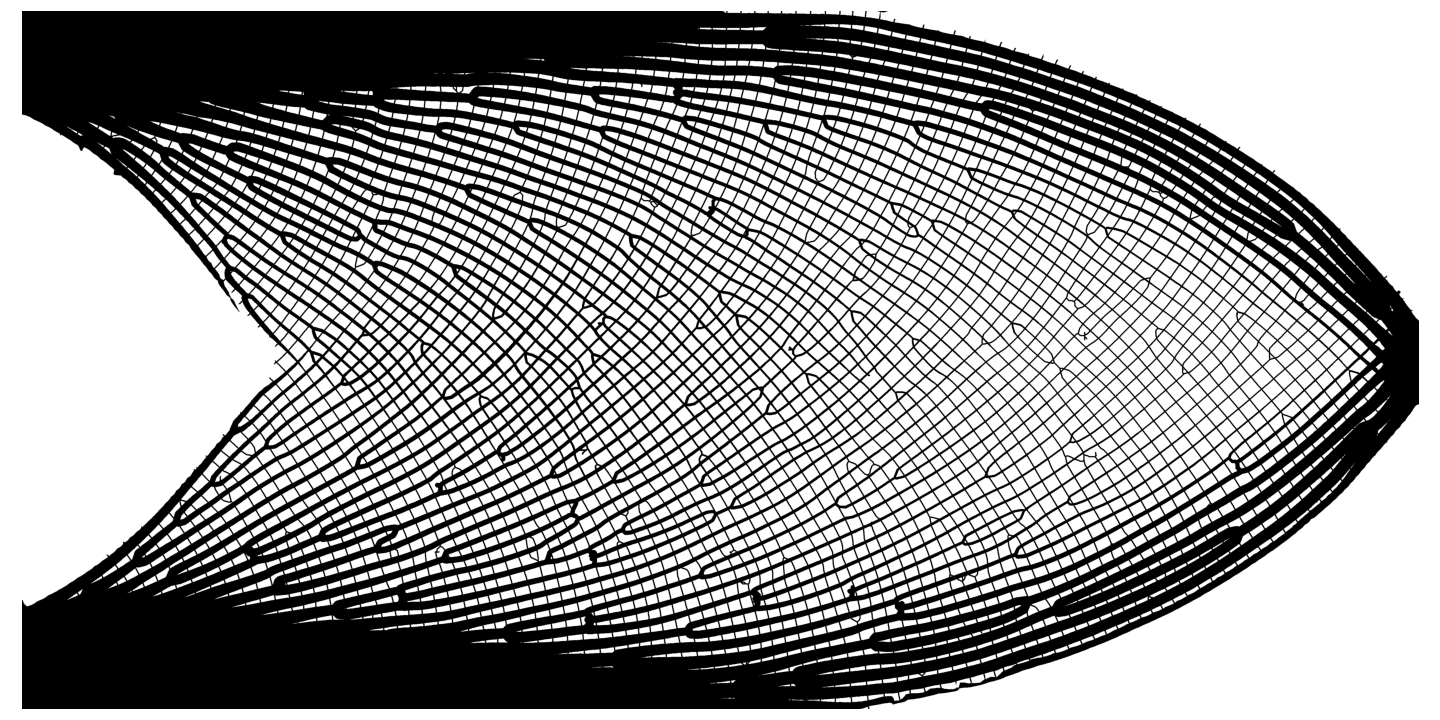}
		\caption{$V=0.40$, $\mu_{min}=0.10$}
		\label{fig:michell_240_1200_vol_0.40_MinMu_0.10_p20}
	\end{subfigure}
	\begin{subfigure}[t]{0.48\textwidth}
		\centering
		\includegraphics[width=\textwidth]{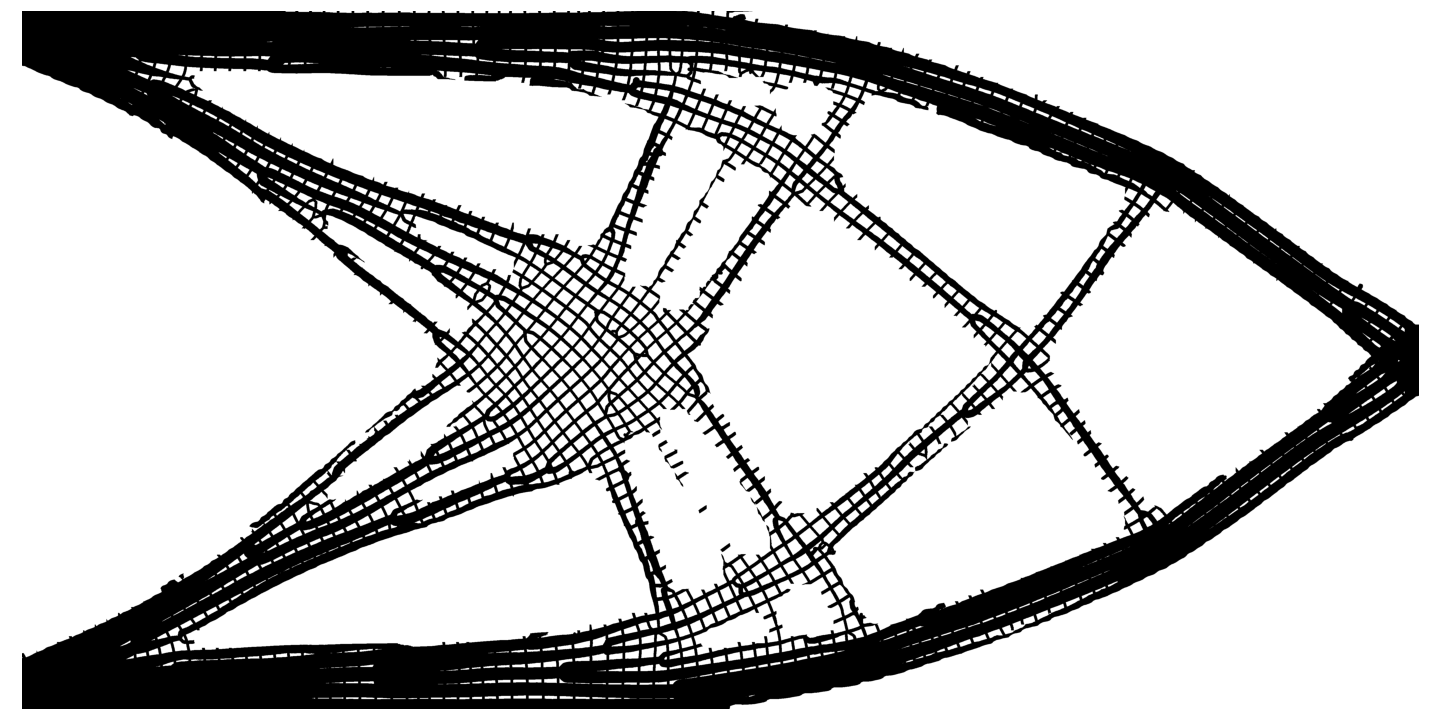}
		\caption{$V=0.25$, $\mu_{min}=0.20$}
		\label{fig:michell_240_120_vol_0.25_MinMu_0.20_p20}
	\end{subfigure}
	\hfill
	\begin{subfigure}[t]{0.48\textwidth}
		\centering
		\includegraphics[width=\textwidth]{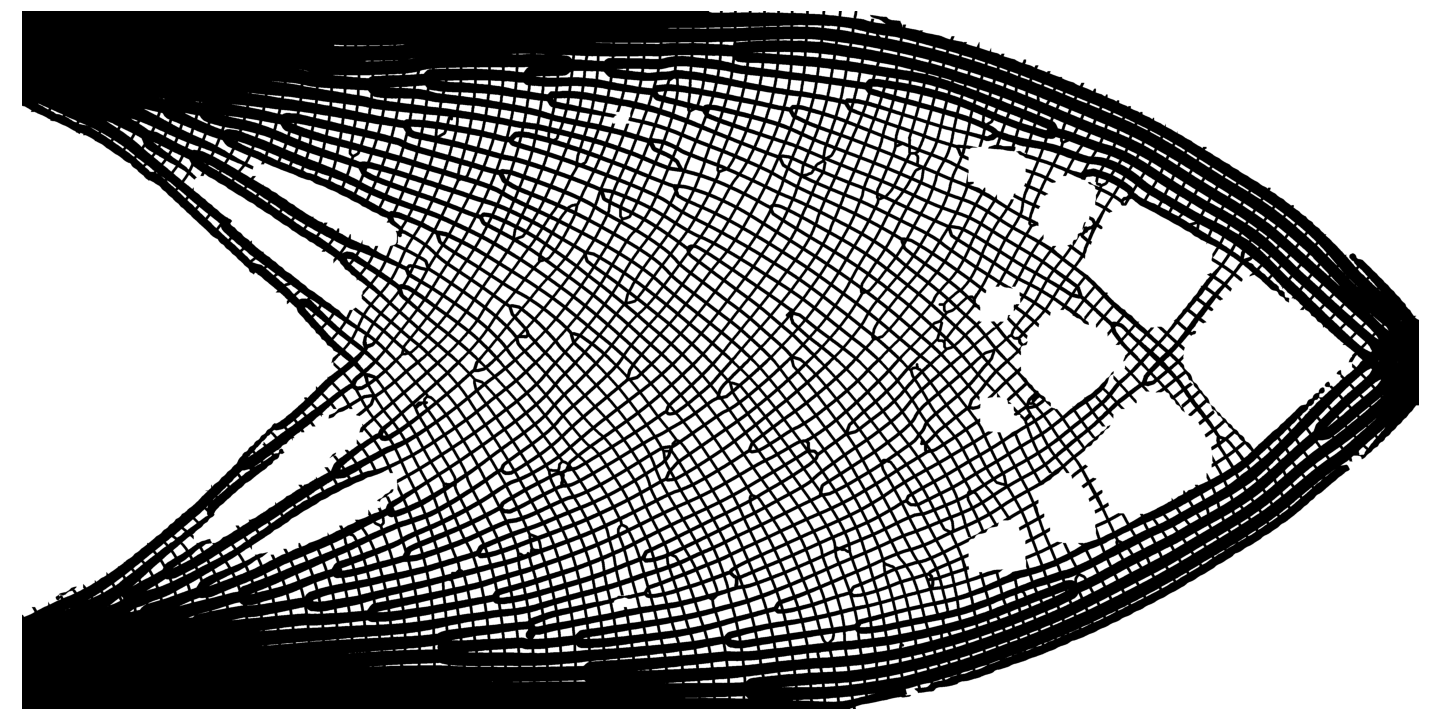}
		\caption{$V=0.40$, $\mu_{min}=0.20$}
		\label{fig:michell_240_120_vol_0.40_MinMu_0.20_p20}
	\end{subfigure}
	\caption{De-homogenization of a $240\times120$ Michell cantilever input to a fine mesh of $5760\times2880$ elements with a wave-length of $\varepsilon_f=60h_f$.}
	\label{fig:michell_high_res_p20}
\end{figure}

\subsubsection{Computational cost}

The computational costs associated with the method are much lower than the standard TO approach, and outperforms current state-of-the-art de-homogenization methods by several factors. The de-homogenization of the $60\times30$ homogenization-based solution onto $1440\times720$ and $2400\times1200$ meshes took $\approx1.3\mathrm{s}$ and $\approx2.2\mathrm{s}$ respectively. At the same resolutions the \texttt{top88} code\footnote{\url{https://www.topopt.mek.dtu.dk/Apps-and-software/Efficient-topology-optimization-in-MATLAB}} took roughly $16\mathrm{s}$ and $87\mathrm{s}$ seconds per iteration respectively. To obtain a design of similar discretization and quality, a continuation strategy would be needed, which typically requires around 1000 iterations to converge \cite{sigmund2016a} leading to a total computation time of 16.000 and 87.000 seconds respectively. Thus, the neural network-based de-homogenization approach is four orders of magnitudes faster.\\
In comparison to other de-homogenization approaches; Groen et al. \cite{groen2018a} report a computation time of roughly 10 seconds when solving the least-squares problem on an intermediate mesh to project a $80\times40$ homogenization-based solution to a $1600\times800$ mesh. Due to other hardware platform, the computation times are not directly comparable, but both set of computations have been carried out on a modern day laptop, and thus we estimate a decrease in computational cost of around a factor 5 to 10.\\

\begin{figure}[H]
	\centering
	\includegraphics[width=0.7\textwidth]{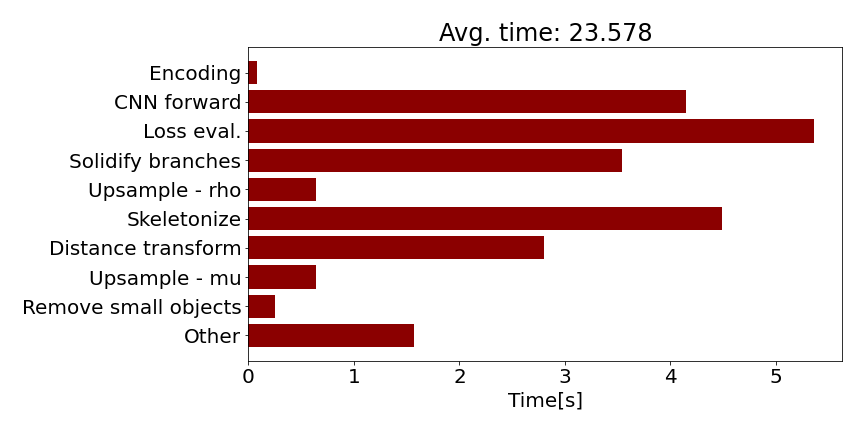}
	\caption{Average computation time for different sub-components of the de-homogenization framework when run on six high-resolution ($5760\times2880$) examples.}
	\label{fig:comp_times_5760x2880}
\end{figure}

A nice property of the proposed method is that the computational cost of all major operations (convolution, bi-linear interpolation, skeletonization and euclidean distance transform) scale linearly in time, making the method very well-suited for generating very high-resolution designs. As an example, for the $5760\times2880$ mesh (16.5 million elements) direct solvers either become extremely memory consuming or break down entirely, and parallel computing \cite{aage2015a} in combination with e.g. multi-grid \cite{amir2014a} preconditioners would have to be leveraged in order to perform standard TO. With the proposed method the optimization could be carried out on a coarse scale mesh, e.g. $240\times120$ elements, using the homogenization method, and afterwards projected onto the high-resolution mesh. All of this could be accomplished in a couple of minutes on a modern day laptop without having to leverage complicated multi-processing frameworks or multi-scale solvers. Figure \ref{fig:comp_times_5760x2880} shows the average computation time for the six designs in Figure \ref{fig:michell_high_res_p20}, along with a breakdown of the computation time for different sub-components of the framework. Here the loss function evaluation, solidify branches, network forward pass, skeletonize and distance transform are the most costly components. The solidify branches component is mostly a safety measure, to make sure that branches are always solid, but is often not needed as the two-step training procedure is enough to assure this in most cases. As this is the only component in the post-processing procedure relying on a loss function evaluation dropping this component could save up to $40\%$ of the computational cost. The three remaining costly components can, as mentioned earlier, all be run on a GPU for a significant speed-up. In terms of memory consumption the proposed method is able to handle inputs up to a resolution of $800\times800$ on a single 11GB GPU. This corresponds to a de-homogenized design with a resolution of $32000\times32000$ when using $\varepsilon_i=10$ which should be more than enough for representing any currently manufacturable design in 2D. For 3D purposes the framework can with no further theoretical development be extended to multi-GPU usage by wrapping the model with a distribution module, such as \texttt{tf.distribute}\footnote{\url{https://www.tensorflow.org/api_docs/python/tf/distribute}} in Tensorflow or \texttt{torch.nn.DataParallel}\footnote{\url{https://pytorch.org/docs/stable/generated/torch.nn.DataParallel.html}} in PyTorch.\\
Considering these significant advantages in terms of computational cost a loss in performance of $7-25\%$ is not immense, and in the low-resolution case where the de-homogenization only takes a few seconds the proposed approach could serve as a real-time evaluation tool for a discrete and manufacturable version of the homogenized-based design. In the high-resolution case the de-homogenized design could serve as a good starting guess for high-resolution shape or topology optimization, thus saving a large amount of iterations.

\subsection{Double-clamped beam}

As a second example, the proposed method is applied to the double-clamped beam with center loading. An input of size $200\times50$ is considered and de-homogenized at two different wave-lengths with the fine-scale mesh size chosen according to the minimum relative thickness. The results are shown in Table \ref{tab:double_clamped_results}. Here the average performance is $22.4\%$ worse than the reference case, and the volume constraint is violated by an average of $2.3\%$. The best performance is achieved for the small wave-length case, but at a cost of double the computation time.

\rowcolors{2}{gray!25}{white}
\begin{table}[H]
	\centering
	\begin{tabular}{c c c | c c | c c c c c | c}
		\rowcolor{gray!50}
		$h_c$ & $\varepsilon_i$ &  $\mu_{min}$ & $V_{ref}$ & $\mathcal{C}_{ref}$ & $h_f$ & $\varepsilon_f$ &  $V_f$ & $\mathcal{C}_f$ & $\frac{\mathcal{C}_{f} \cdot V_{f}}{\mathcal{C}_{ref} \cdot V_{ref}}$ & $t_f$[s] \\ \hline
		1/50 & $20h_i$  & 0.05 & 0.2510 & 25.89 & $1/24 h_c$ & $60 h_f$& 0.2590 & 31.35 & 1.2492 & 7.69 \\
		1/50 & $20h_i$  & 0.10 & 0.2538 & 27.11 & $1/24 h_c$ & $60 h_f$& 0.2619 & 32.51 & 1.2369 & 7.12 \\
		1/50 & $10h_i$  & 0.05 & 0.2510 & 25.89 & $1/40 h_c$ & $50 h_f$& 0.2568 & 31.50 & 1.2446 & 15.80 \\	
		1/50 & $10h_i$  & 0.10 & 0.2538 &  27.11 & $1/40 h_c$ & $50 h_f$& 0.2574 & 31.12 & 1.1657 & 14.70 \\
	\end{tabular}
	\caption{Performance and computational cost of the neural network based de-homogenization approach on the double-clamped beam example.}
	\label{tab:double_clamped_results}
\end{table}
\rowcolors{2}{white}{white}

\begin{figure}[H]
	\centering
	\begin{subfigure}[t]{\textwidth}
		\centering
		\includegraphics[width=0.8\textwidth]{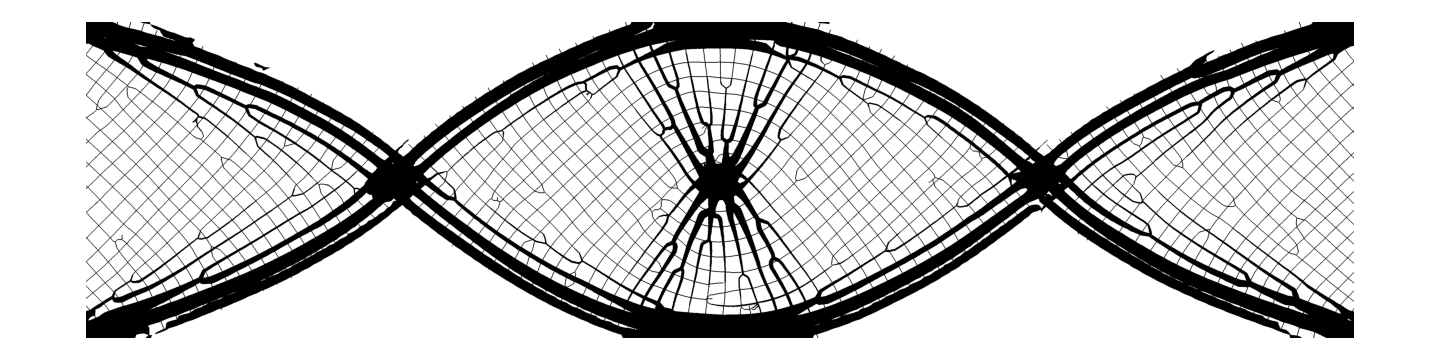}
		\caption{$V=0.25$, $\mu_{min}=0.05$, $\varepsilon_f=60h_f$}
		\label{fig:double_clamped_200_50_vol_0.25_MinMu_0.05_p20}
	\end{subfigure}
	\begin{subfigure}[t]{\textwidth}
		\centering
		\includegraphics[width=0.8\textwidth]{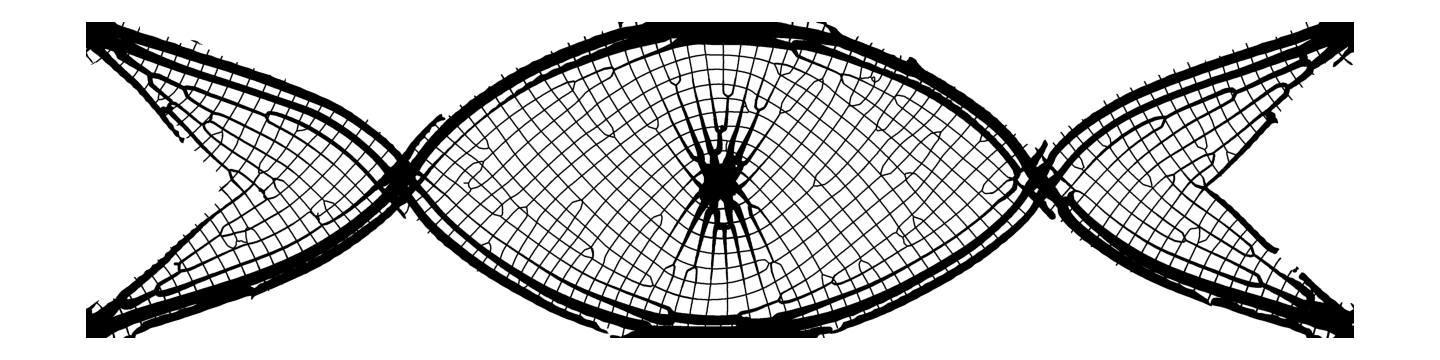}
		\caption{$V=0.25$, $\mu_{min}=0.10$, $\varepsilon_f=60h_f$}
		\label{fig:double_clamped_200_50_vol_0.25_MinMu_0.10_p20}
	\end{subfigure}
	\begin{subfigure}[t]{\textwidth}
		\centering
		\includegraphics[width=0.8\textwidth]{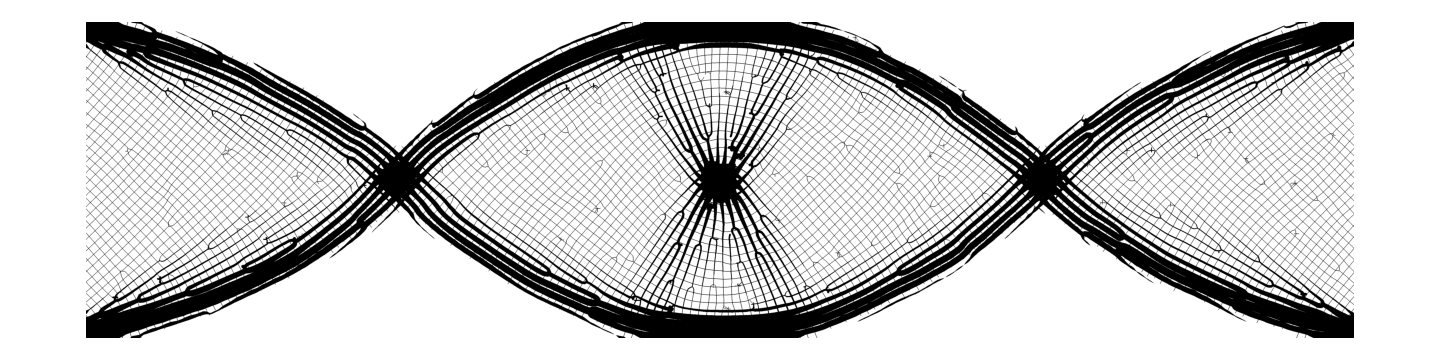}
		\caption{$V=0.25$, $\mu_{min}=0.05$, $\varepsilon_f=50h_f$}
		\label{fig:double_clamped_200_50_vol_0.25_MinMu_0.05_p10}
	\end{subfigure}
	\begin{subfigure}[t]{\textwidth}
		\centering
		\includegraphics[width=0.8\textwidth]{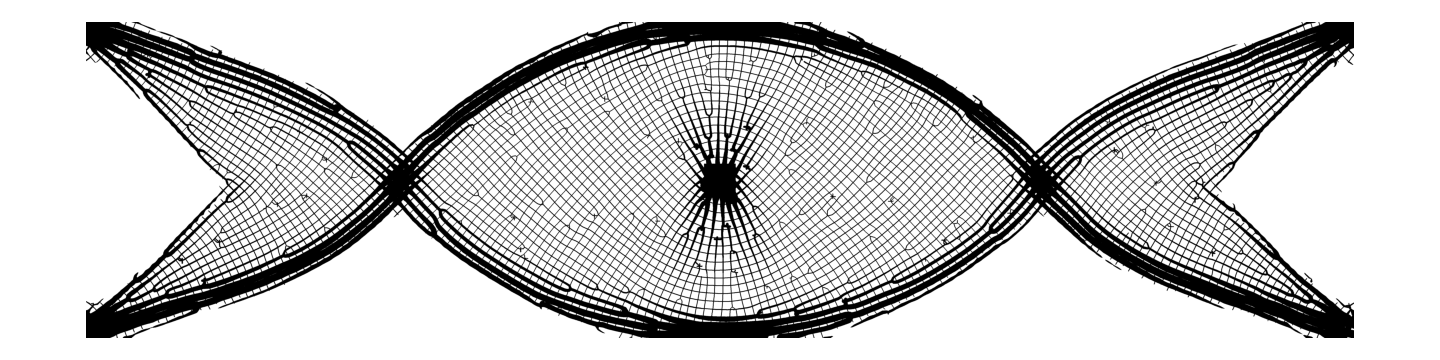}
		\caption{$V=0.25$, $\mu_{min}=0.10$, $\varepsilon_f=50h_f$}
		\label{fig:double_clamped_200_50_vol_0.25_MinMu_0.10_p10}
	\end{subfigure}
	\caption{De-homogenization of a $200\times50$ double-clamped beam input to a fine mesh of $4800\times1200$ (\textbf{a} and \textbf{b}) and $8000\times2000$ (\textbf{c} and \textbf{d}).}
	\label{fig:double_clamped}
\end{figure}

\subsection{L-shaped beam}

\begin{figure}[H]
	\centering
	\begin{subfigure}[t]{0.40\textwidth}
		\centering
		\includegraphics[width=\textwidth]{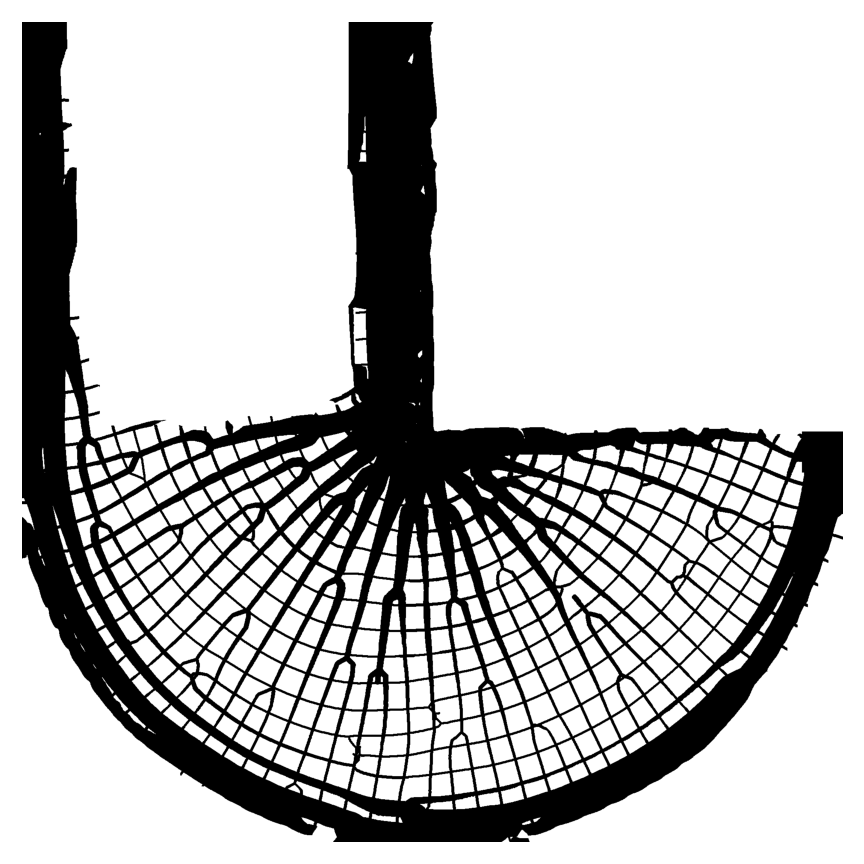}
		\caption{$1920\times1920$ mesh, $\varepsilon_f = 60h_f$}
		\label{fig:l_shaped_beam_80_80_combed_vol_0.25_MinMu_0.10_p20}
	\end{subfigure}
	\hfill
	\begin{subfigure}[t]{0.40\textwidth}
		\centering
		\includegraphics[width=\textwidth]{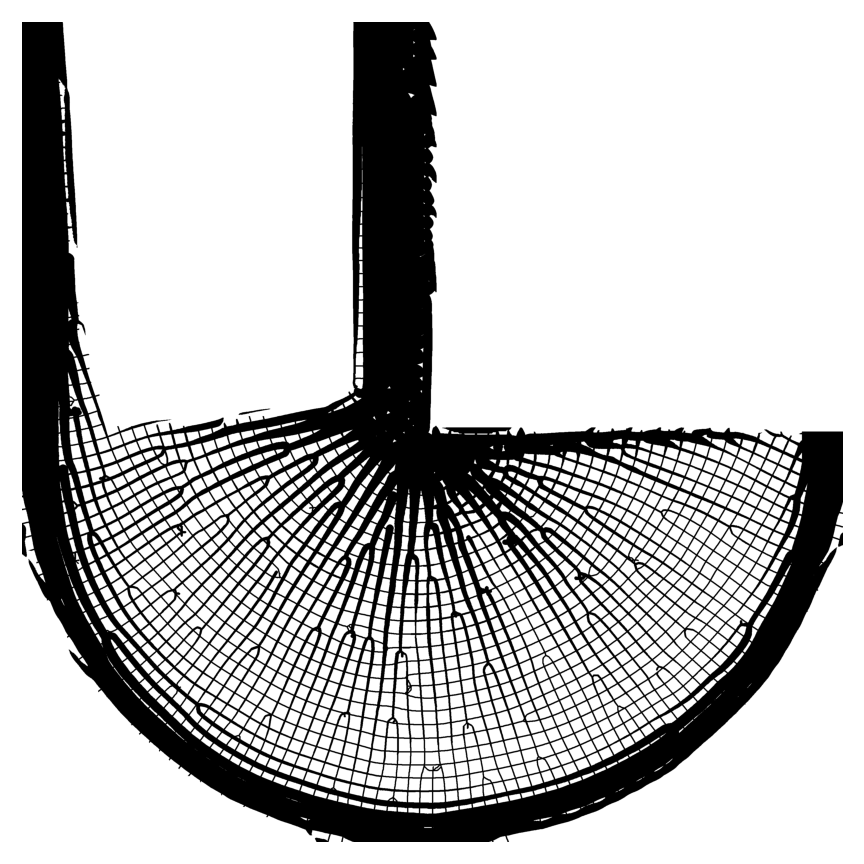}
		\caption{$3200\times3200$ mesh, $\varepsilon_f = 50h_f$}
		\label{fig:l_shaped_beam_80_80_combed_vol_0.25_MinMu_0.10_p10}
	\end{subfigure}
	\caption{De-homogenization of the L-shaped beam with a combed orientation field as input to the neural network. $V=0.25$, $\mu_{min}=0.10$.}
	\label{fig:l_shaped_combed}
\end{figure}

As a last example the L-shaped beam is considered. This example presents a more challenging case due to the singularity at the sharp corner in the design domain. Directly solving the optimization problem in eq. \ref{eq:opt_problem} leads to small regions near the sharp corner, in which the angles are turned $90^{\circ}$ compared to neighboring elements. One way to remedy this issue is the vector field combing method presented in Stutz et al. \cite{stutz2020a} which provides a consistent labeling of the lamination orientations. Table \ref{tab:l_shaped_beam_results} and Figure \ref{fig:l_shaped_combed} shows the results for the L-shaped beam using the combed input fields. Here it can be seen that higher periodicity only leads to a slightly better performance with both designs being within roughly $12\%$ of the reference solution.

\rowcolors{2}{gray!25}{white}
\begin{table}[H]
	\centering
	\begin{tabular}{c c c | c c | c c c c c | c}
		\rowcolor{gray!50}
		$h_c$ & $\varepsilon_i$ &  $\mu_{min}$ & $V_{ref}$ & $\mathcal{C}_{ref}$ & $h_f$ & $\varepsilon_f$ & $V_f$ & $C_f$ & $\frac{\mathcal{C}_{f} \cdot V_{f}}{\mathcal{C}_{ref} \cdot V_{ref}}$ & $t_f$[s] \\ \hline
		1/80 & $20h_i$  & 0.10  & 0.2517  & 94.15   & $1/24h_c$ & $60h_f$ & 0.2758 & 96.29 & 1.1207 & 4.96\\
		1/80 & $10h_i$  & 0.10 & 0.2517  & 94.15   & $1/40 h_c$ & $50h_f$ & 0.2742 & 96.52 & 1.1168 & 7.53 \\
	\end{tabular}
	\caption{Performance and computational cost of neural network based de-homogenization approach on the L-shaped beam.}
	\label{tab:l_shaped_beam_results}
\end{table}

\section{Conclusions and future work} \label{sec:conclusion}

An unsupervised deep learning-based de-homogenization framework capable of generating high-resolution, manufacturable, and continuous microstructures has been presented. Compared to previous deep learning-based topology optimization approaches the proposed method does not seek to perform topology optimization in an end-to-end manner, but instead utilizes the solution from a traditional homogenization-based topology optimization as input. 
This leads to the, perhaps, most important feature of the proposed methodology: It is, to a large extent, insensitive to domain size, loading, and boundary conditions. The proposed fully convolutional neural network approach is capable of synthesizing continuous and periodic microstructures, while the input, in the form of an orientation field, guides the network towards a mechanically sound design. A subsequent post-processing scheme assigns an appropriate thickness to each member of the final design, and ensures manufacturability by imposing a minimum relative thickness. Training of the neural network is extremely inexpensive as synthetic orientation fields can be used as training data, and generalizes to a wide range of problems, as the underlying load and boundary conditions are implicitly enforced by the homogenization-based input. Numerical examples show that the proposed method is a factor 5 to 10 faster than current state-of-the-art de-homogenization approaches and has a performance within $7-25\%$ of the reference solution (before post-processing). While only 2D examples are shown the very low computational cost of the method makes it an obvious candidate for 3D applications. Furthermore, as all operations in the proposed de-homogenization pipeline support GPU computations, a GPU implementation could pave the way for interactive de-homogenization, i.e. projection during the optimization, in 2D cases. \\

There are several ways upon which the current method could be improved. For example, lamination widths could be used directly as input to the neural network, instead of being applied in the post-processing scheme. This would allow the neural network to trade a high error in inconsequential parts of the domain, i.e. regions with very little material, for a lower error in solid parts of the domain. As of now, the proposed method also tends to slightly overshoot the prescribed volume fraction. This problem could be remedied by iteratively performing a finite element analysis to identify non-loaded carrying members in the final design and remove them. Furthermore, a graph-based post-processing scheme could be used to identify U-shaped branches and turn them into V-shaped branches which are known to be a mechanically better design when designing for minimum compliance. Finally, the iso-contour of the design could be extracted directly from the distance transform utilized in the post-processing scheme, such that the design can be described using a conforming mesh. This would yield a smoother boundary representation while allowing a more efficient finite element analysis during post-processing and subsequent shape optimization. An obvious idea, to be studied in future work, is the extension to 3D, which is expected to result in even larger time savings than reported here for the 2D case.

\subsection*{Acknowledgements}

The authors would like to acknowledge the support of the Villum Foundation through the InnoTop project. Gratitude should also be expressed to Anders Nymark Christensen and Morten Hannemose for valuable discussions on the deep learning parts of the project. Finally, the authors would like to thank Jeroen Peter Groen for providing the homogenization-based topology optimization code and to Florian Stutz for providing the code for angular field combing.

\appendix
\section{Michell cantilever - 120x60 designs}
\setcounter{figure}{0}
\setcounter{table}{0}

\rowcolors{2}{gray!25}{white}
\begin{table}[H]
	\centering
	\begin{tabular}{c c c | c c | c c c c c | c}
		\rowcolor{gray!50}
		$h_c$ & $\varepsilon_i$ & $\mu_{min}$ & $V_{ref}$ & $\mathcal{C}_{ref}$ & $h_f$ & $\varepsilon_f$ & $V_f$ & $\mathcal{C}_f$ & $\frac{\mathcal{C}_{f} \cdot V_{f}}{\mathcal{C}_{ref} \cdot V_{ref}}$ & $t_f$[s] \\ \hline
		1/60 & $20h_i$ & 0.05  & 0.2534 & 105.92 & $1/24 h_c$ & $60 h_f$ & 0.2608 & 128.52 & 1.2489 & 5.54 \\
		1/60 & $20h_i$& 0.05 & 0.4024 & 68.58 & $1/24 h_c$& $60 h_f$ & 0.4219 & 74.36 & 1.1366 & 5.52 \\
		1/60 & $20h_i$ & 0.10 & 0.2567 & 112.42 & $1/24 h_c$ & $60 h_f$ & 0.2599 & 134.91 & 1.2148 & 5.85 \\
		1/60 & $20h_i$ & 0.10 & 0.4080 & 68.95 &  $1/24 h_c$& $60 h_f$ & 0.4309 & 71.89 & 1.1010 & 5.38 \\
		1/60 & $20h_i$ & 0.20 & 0.2593 & 117.60 & $1/24 h_c$ & $60 h_f$ & 0.2601 & 175.84 & 1.4998 & 5.61 \\
		1/60 & $20h_i$ & 0.20 & 0.4163 & 72.89 & $1/24 h_c$ & $60 h_f$ & 0.4391 & 74.60 & 1.0798 & 5.45
	\end{tabular}
	\caption{Performance and computational cost of neural network based de-homogenization approach on the Michell cantilever beam for a $120\times60$ input.}
	\label{tab:michell_cantilever_120x60}
\end{table}
\rowcolors{2}{white}{white}

\begin{figure}[H]
	\centering
	\begin{subfigure}[t]{0.48\textwidth}
		\centering
		\includegraphics[width=\textwidth]{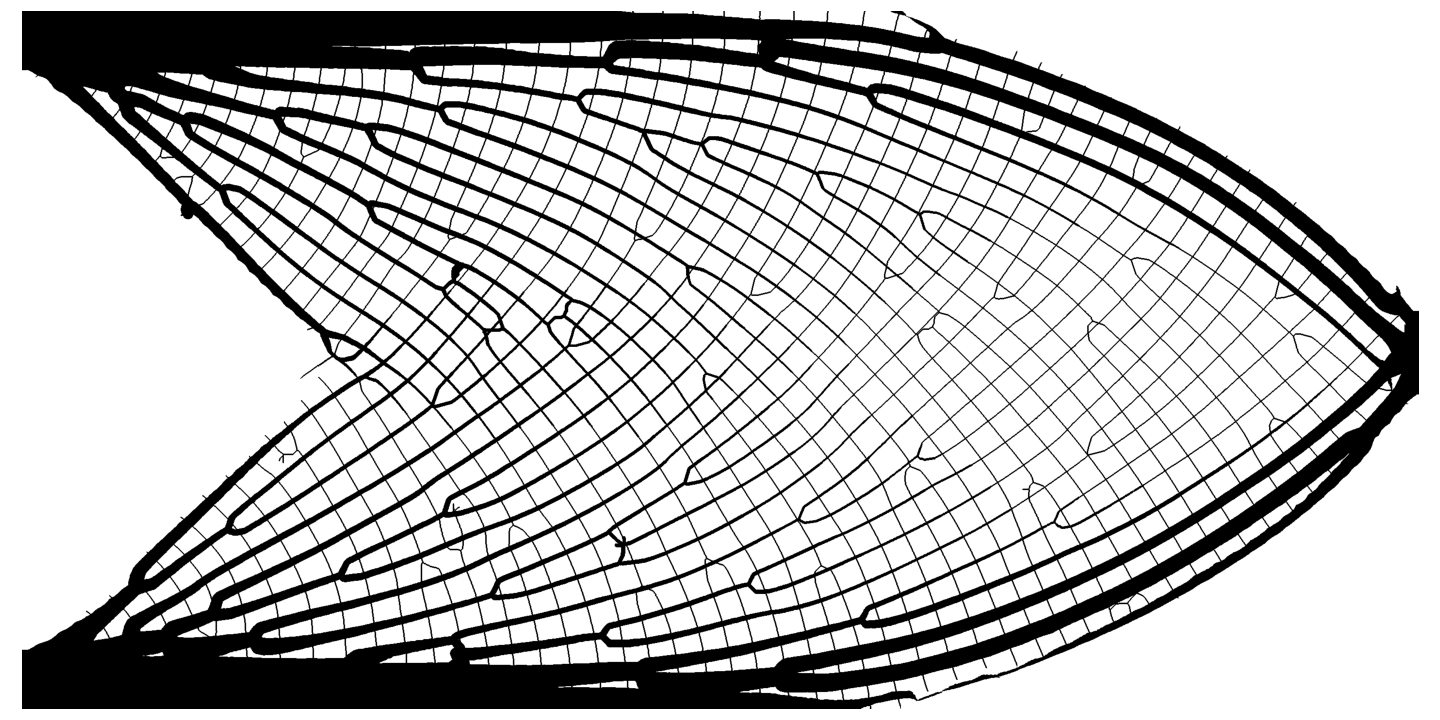}
		\caption{$V=0.25$, $\mu_{min}=0.05$}
	\end{subfigure}
	\hfill
	\begin{subfigure}[t]{0.48\textwidth}
		\centering
		\includegraphics[width=\textwidth]{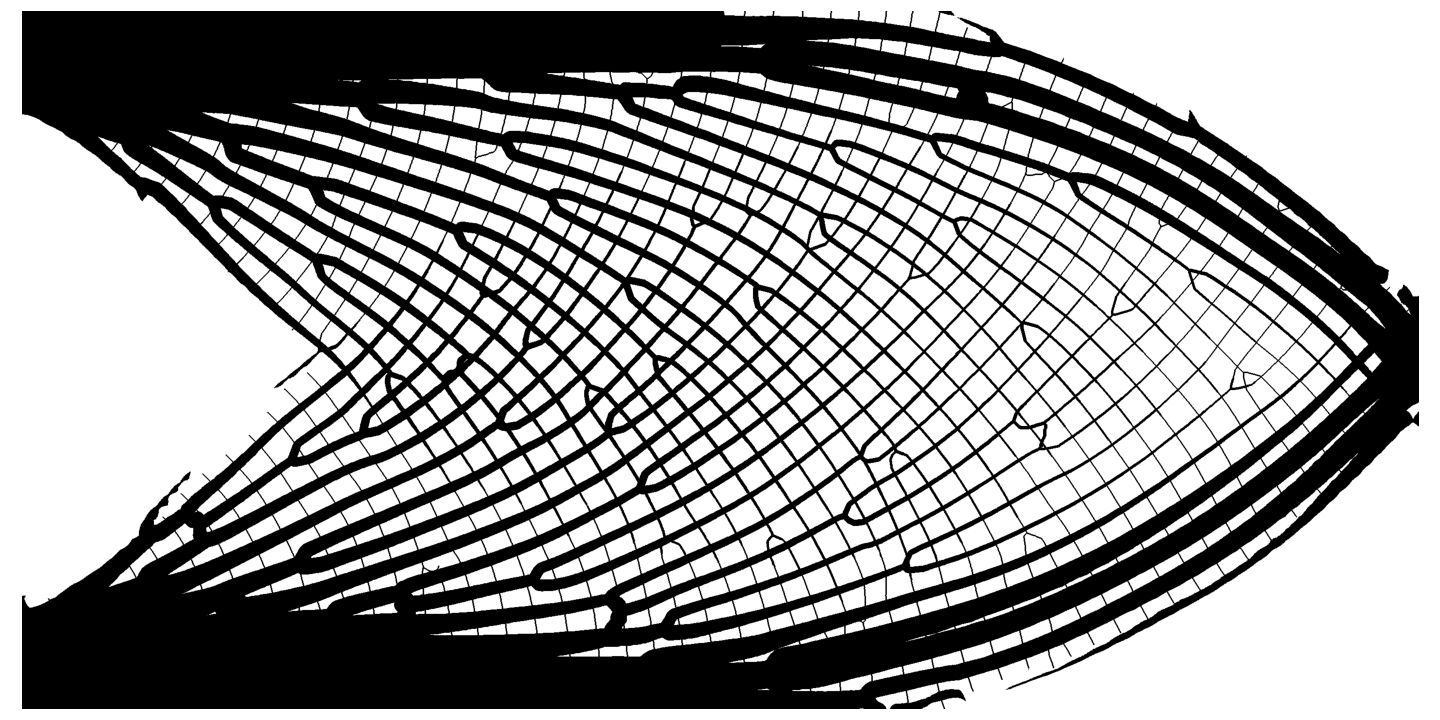}
		\caption{$V=0.40$, $\mu_{min}=0.05$}
	\end{subfigure}
	\begin{subfigure}[t]{0.48\textwidth}
		\centering
		\includegraphics[width=\textwidth]{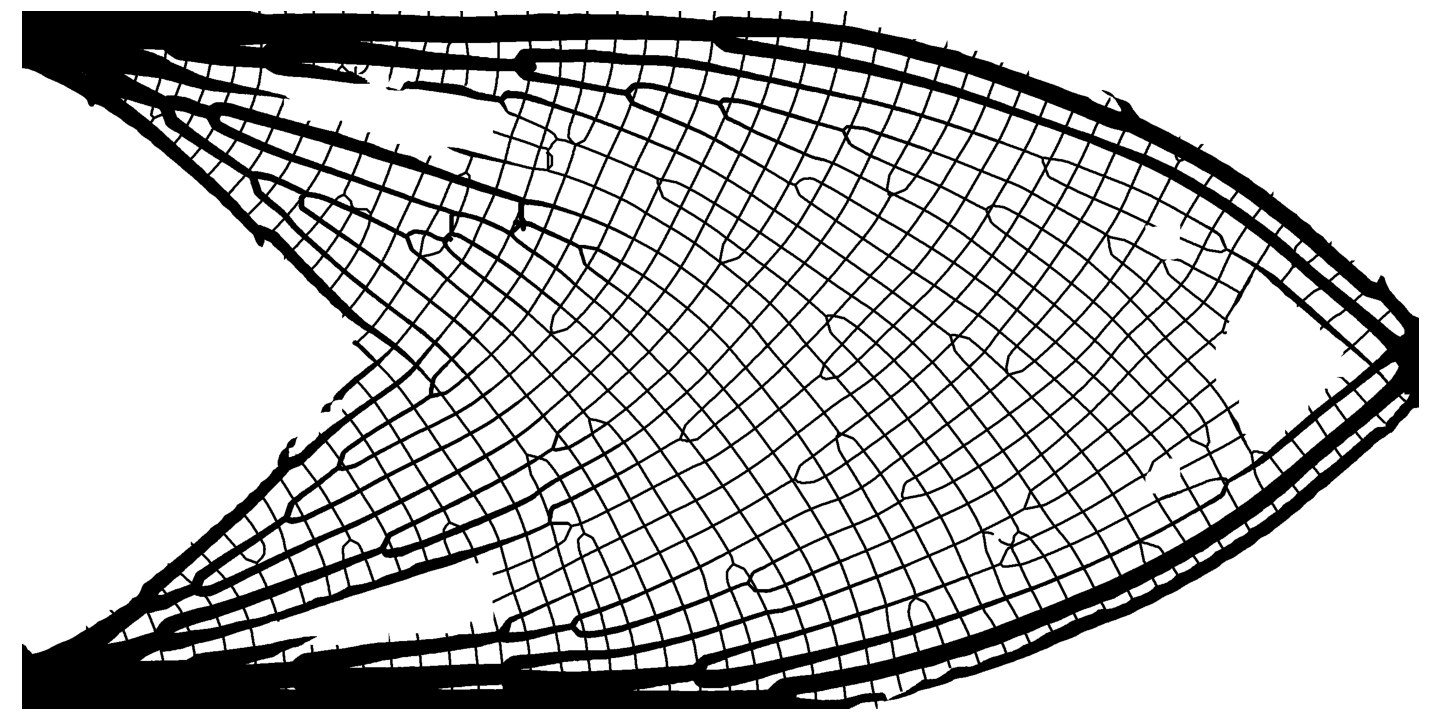}
		\caption{$V=0.25$, $\mu_{min}=0.10$}
	\end{subfigure}
	\hfill
	\begin{subfigure}[t]{0.48\textwidth}
		\centering
		\includegraphics[width=\textwidth]{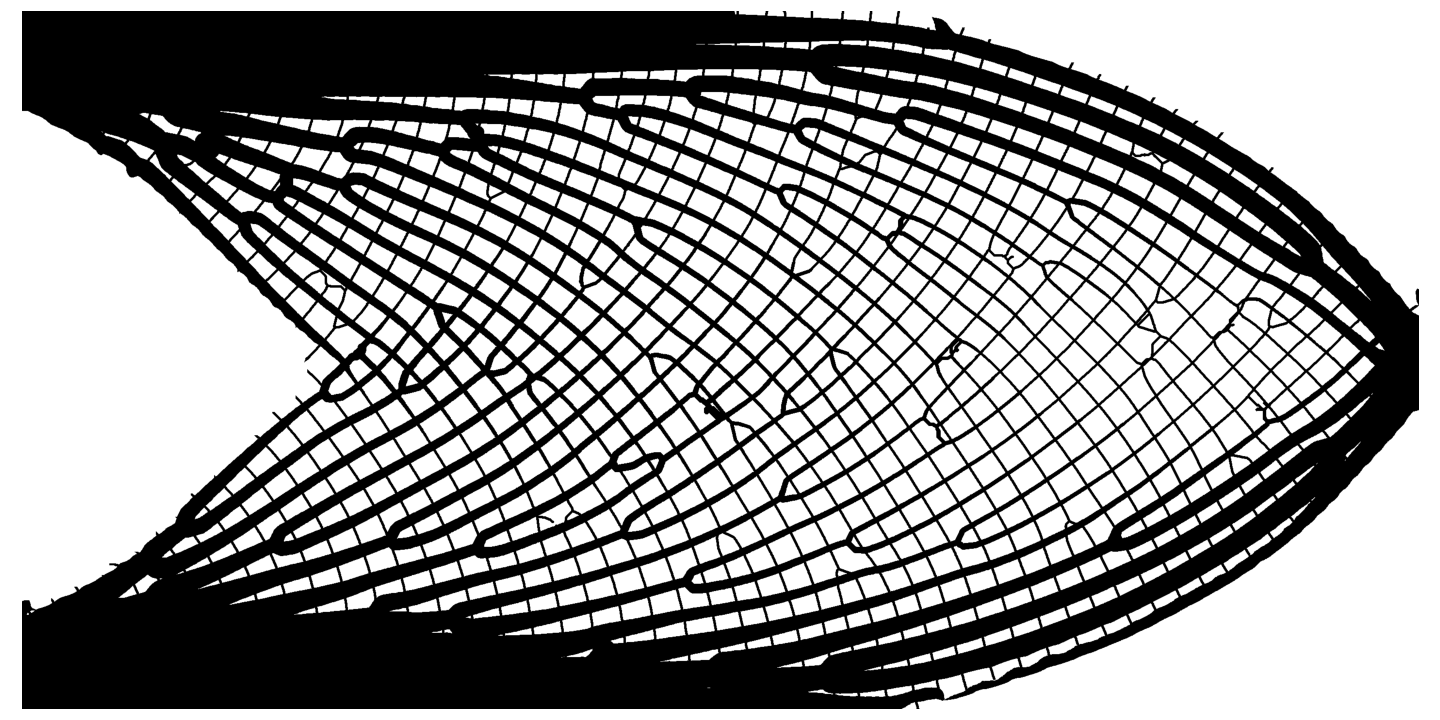}
		\caption{$V=0.40$, $\mu_{min}=0.10$}
	\end{subfigure}
	\begin{subfigure}[t]{0.48\textwidth}
		\centering
		\includegraphics[width=\textwidth]{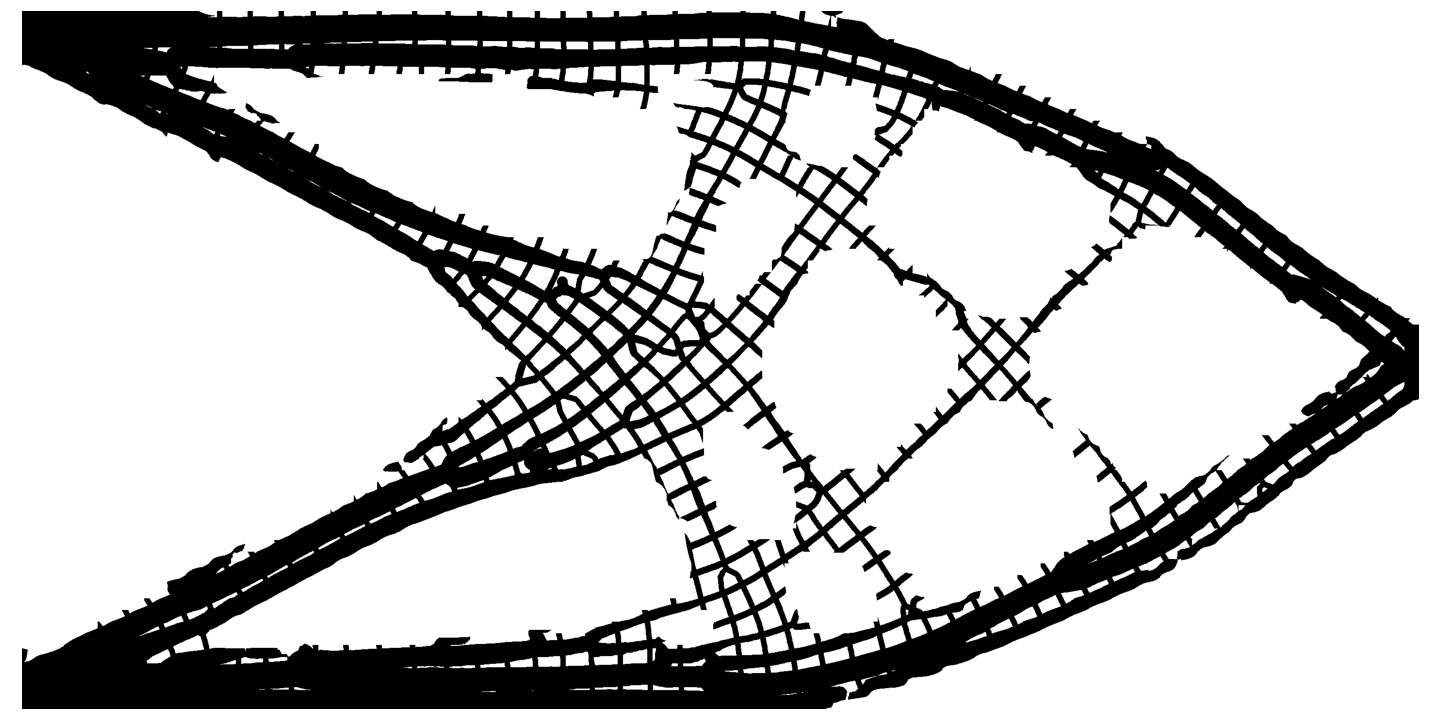}
		\caption{$V=0.25$, $\mu_{min}=0.20$}
	\end{subfigure}
	\hfill
	\begin{subfigure}[t]{0.48\textwidth}
		\centering
		\includegraphics[width=\textwidth]{fig/michell/michell_60_30_vol_040_MinMu_020_p20.png}
		\caption{$V=0.40$, $\mu_{min}=0.20$}
	\end{subfigure}
	\caption{De-homogenization of a $120\times60$ Michell cantilever input to a fine mesh of $2880\times1440$ elements with a wave-length of $\varepsilon_f=60h_f$.}
	\label{fig:michell_120x60_p20}
\end{figure}

\section{Network architecture} \label{app:net_arch}
\setcounter{figure}{0}
\setcounter{table}{0}

The neural network consists of 9 convolutional layers, four ResNet blocks and three upsampling layers. The two first layers use a kernel size of 7 and 5 respectively, while all other layers use a kernel size of 3. Between each convolutional layer batch normalization is applied, and ReLU is used the activation function in all layers except the last, which uses a Sigmoid. For upsampling nearest neighbors is used, as this is less prone to checkerboard artifacts than learned upsampling filters. The network architecture is summarized in Table \ref{tab:network_arch}.

\begin{table}[H]
	\centering
	\begin{tabular}{c | l}
		\textbf{Layer} & \textbf{Description} \\ \hline
		Input & $\mathbb{R}^{N}$ $\mathrm{H}\times \mathrm{W}$ \\ \hline
		1 & $7\times7$ conv, $C=32$, BatchNorm, ReLU \\ \hline
		2 & $5\times5$ conv, $C=64$, BatchNorm, ReLU \\ \hline
		3 & ($3\times3$ conv, $C=64$, ResNet block) $\times$ 4 \\ \hline
		4 & Nearest neighbour upsampling, scale=2 \\ \hline
		5 & $3\times3$ conv, $C=64$, BatchNorm, ReLU \\ \hline
		6 & $3\times3$ conv, $C=64$, BatchNorm, ReLU \\ \hline
		7 & Nearest neighbour upsampling, scale=2 \\ \hline
		8 & $3\times3$ conv, $C=64$, BatchNorm, ReLU \\ \hline
		9 & $3\times3$ conv, $C=32$, BatchNorm, ReLU \\ \hline
		10 & Nearest neighbour upsampling, scale=2 \\ \hline
		11 & $3\times3$ conv, $C=32$, BatchNorm, ReLU \\ \hline
		12 & $3\times3$ conv, $C=32$, BatchNorm, ReLU \\ \hline
		13 & $3\times3$ conv, $C=1$, Sigmoid
	\end{tabular}
	\caption{Convolutional neural network architecture}
	\label{tab:network_arch}
\end{table}

\bibliography{references}

\end{document}